\documentclass[times,twocolumn,final]{elsarticle}

\usepackage{medima}
\usepackage{framed,multirow}

\usepackage{latexsym}

\usepackage{amsmath}
\usepackage{multirow}
\usepackage{longtable}
\usepackage{amssymb}
\usepackage{lipsum}
\usepackage{subfigure}
\usepackage{graphicx}
\usepackage{pifont}
\usepackage{amsmath}
\usepackage{amsthm}
\usepackage{makecell}
\usepackage{amsmath,amssymb,amsfonts}
\usepackage{algorithm}
\usepackage{algorithmic}
\usepackage{graphicx}
\usepackage{textcomp}

\usepackage{microtype}
\usepackage{multirow}
\usepackage{booktabs}
\usepackage{rotating}
\usepackage{soul}
\usepackage{cancel}

\usepackage{url}

\usepackage{hyperref}
\usepackage[dvipsnames, table]{xcolor}
\usepackage{caption}
\journal{Medical Image Analysis}

\begin{document}

\verso{F. Tang, B. Kun, Y. Li \textit{et~al.}}

\begin{frontmatter}

\title{MambaMIM: Pre-training Mamba with State Space Token Interpolation and its Application to Medical Image Segmentation}

\author[1,2,3]{Fenghe \snm{Tang}\fnref{fn1}}

\author[4,5,6]{Bingkun \snm{Nian}\fnref{fn1}}
\author[1,2,3]{Yingtai \snm{Li}\fnref{fn1}}
\author[1,2,3]{Zihang \snm{Jiang}}
\author[4,5,6]{Jie \snm{Yang}}
\author[4,5,6]{Wei \snm{Liu}\corref{cor1}}
\author[1,2,3,7]{S Kevin \snm{Zhou}\corref{cor1}}

\address[1]{School of Biomedical Engineering, Division of Life Sciences and Medicine, University of Science and Technology of China, Hefei, Anhui, 230026, P.R. China}

\address[2]{Suzhou Institute for Advanced Research, University of Science and Technology of China, Suzhou, Jiangsu, 215123, P.R. China}

\address[3]{Center for Medical Imaging, Robotics, and Analytic Computing \& LEarning (MIRACLE), Suzhou Institute for Advanced Research, USTC, Suzhou, P.R. China}

\address[4]{Department of Automation, Shanghai Jiao Tong University}

\address[5]{Institute of Image Processing and Pattern Recognition, Shanghai Jiao Tong University}

\address[6]{Institute of Medical Robotics, Shanghai Jiao Tong University}

\address[7]{State Key Laboratory of Precision and Intelligent Chemistry, University of Science and Technology of China, Hefei, Anhui 230026, China}

\cortext[cor1]{Corresponding author: s.kevin.zhou@ustc.edu.cn;
  weiliucv@sjtu.edu.cn.}
\fntext[fn1]{These authors contributed equally.}


\begin{abstract}
Recently, the state space model Mamba has demonstrated efficient long-sequence modeling capabilities, particularly for addressing long-sequence visual tasks in 3D medical imaging. However, existing generative self-supervised learning methods have not yet fully unleashed Mamba's potential for handling long-range dependencies because they overlook the inherent causal properties of state space sequences in masked modeling. To address this challenge, we propose a general-purpose pre-training framework called MambaMIM, a masked image modeling method based on a novel {\bf TOKen-Interpolation} strategy (TOKI) for the selective structure state space sequence, which learns causal relationships of state space within the masked sequence. Further, MambaMIM introduces a bottom-up 3D hybrid masking strategy to maintain a {\bf masking consistency} across different architectures and can be used on any single or hybrid Mamba architecture to enhance its multi-scale and long-range representation capability. We pre-train MambaMIM on a large-scale dataset of 6.8K CT scans and evaluate its performance across eight public medical segmentation benchmarks. Extensive downstream experiments reveal the feasibility and advancement of using Mamba for medical image pre-training. In particular, when we apply the MambaMIM to a customized architecture that hybridizes MedNeXt and Vision Mamba, we consistently obtain the state-of-the-art segmentation performance. The code is available at: \url{https://github.com/FengheTan9/MambaMIM}.

\end{abstract}


\end{frontmatter}



\section{Introduction}
\label{sec:introduction}

The advancement in integrating medical image analysis (MIA) into clinical workflows highlights its critical role in modern healthcare, particularly in medical image segmentation and disease diagnosis~\citep{seg_application, deepinmia}. Recently, state space models, especially Mamba~\citep{mamba, mamba2}, have demonstrated promising performance in long-sequence modeling, challenging the widely used Transformer architectures~\citep{transformer}. Vision Mamba~\citep{vim} has emerged as a potential solution for long-range visual modeling in medical imaging, demonstrating competitive results across various downstream tasks~\citep{umamba, segmamba, swinumamba, reconstruction, registration, cls}.
However, the scarcity of labeled data in MIA~\citep{ssl3d, abd1k, word} and the inherent properties of state space sequences~\citep{mamba2} pose significant challenges for training high-performance Vision Mamba models. To address these limitations, researchers have developed hybrid architectures that combine Mamba with convolutional neural networks (CNNs)~\citep{swinumamba, segmamba, umamba}. Despite these advances, developing effective models for MIA on limited labeled data remains a critical challenge.

\begin{figure}
    \centering
    \includegraphics[width=1\linewidth]{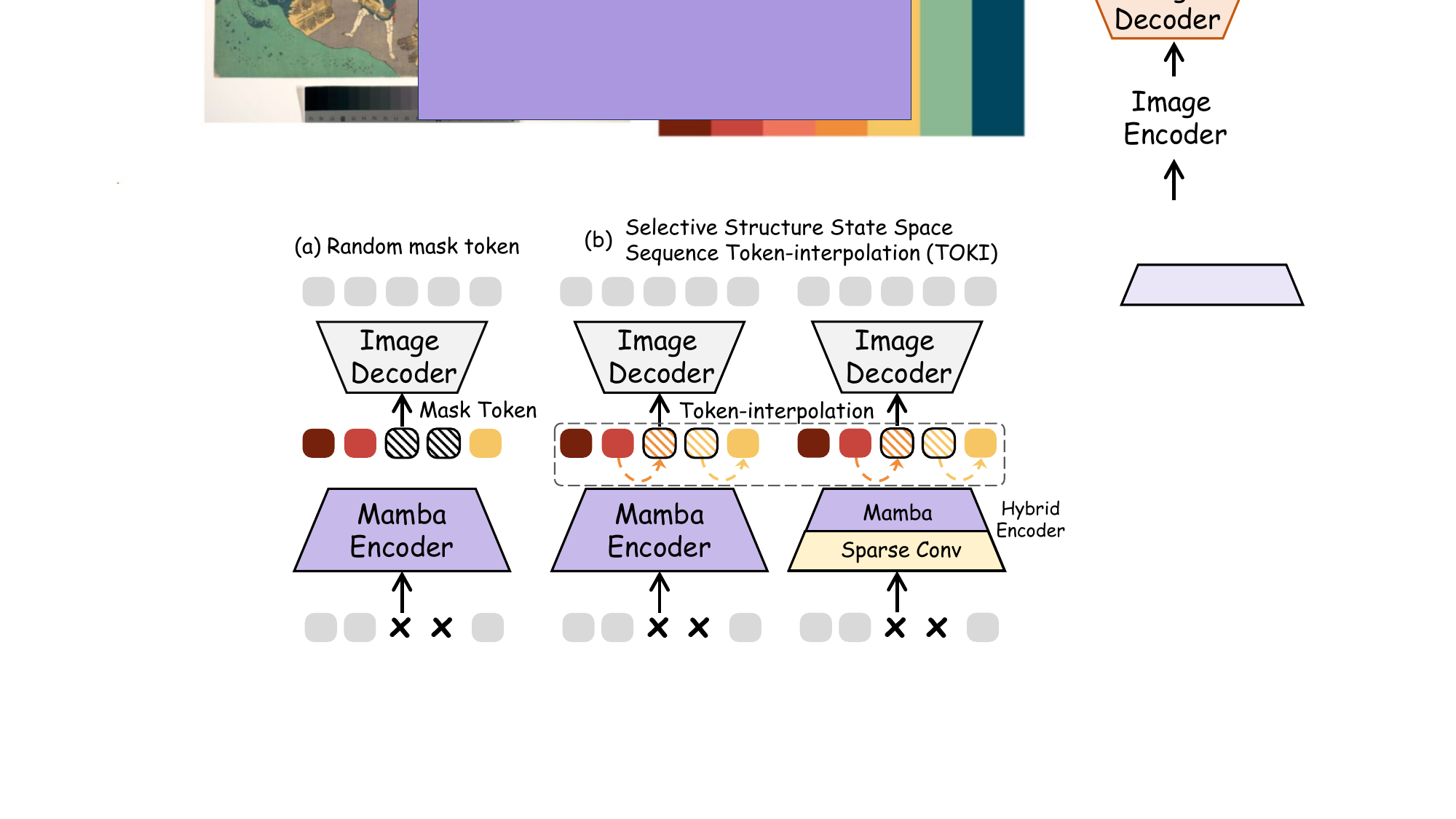}
    \caption{Different token generation strategies for Mamba-based network. (a) Random learnable mask token is generated for decoding. 
    (b) Token-interpolation (TOKI) applies the structure sequence relationships within the state space for Mamba.
    }
    \vspace{-4mm}
    \label{fig:pipeline}
\end{figure}

Fortunately, self-supervised learning (SSL) provides promising solutions~\citep{sslmia}, which first pre-trains a model on a large corpus of easily-collected unlabeled medical images and then transfers it to downstream tasks~\citep{sslmia, mim, survey}. Such a pretrain-finetune paradigm could greatly boost performance in various downstream tasks. Most successful self-supervised learning methods for vision tasks can be categorized as contrastive~\citep{simclr,moco,byol,swav,simsiam,dino} and generative~\citep{bert,inpaint,beit,ibot,mae,simmim,jepa,cmae,spark} methods. Among them, generative methods, represented by a series of 
masked image modeling (MIM) approaches, 
exhibit a stronger transfer ability~\citep{mae,simmim,jepa,cmae,spark} in fine-grand downstream tasks, such as segmentation. However, most current MIM methods focus solely on the contextual relationships between the visible tokens, overlooking how to better modeling the inherent causal properties of state-space sequences. In addition, MIM methods necessitate handling both masked and seen patches, typically relying on strategies specific to certain architectures~\citep{simmim, mae, spark}, \textit{e.g.} MAE~\citep{mae} for ViT~\citep{vit} and SparK~\citep{spark} for CNN~\citep{unet}. Adapting these approaches to single and hybrid Vision Mamba is particularly challenging, especially for 3D medical images that demand modeling significantly longer sequences.

\begin{figure}[h]
    \centering
    \includegraphics[width=0.9\linewidth]{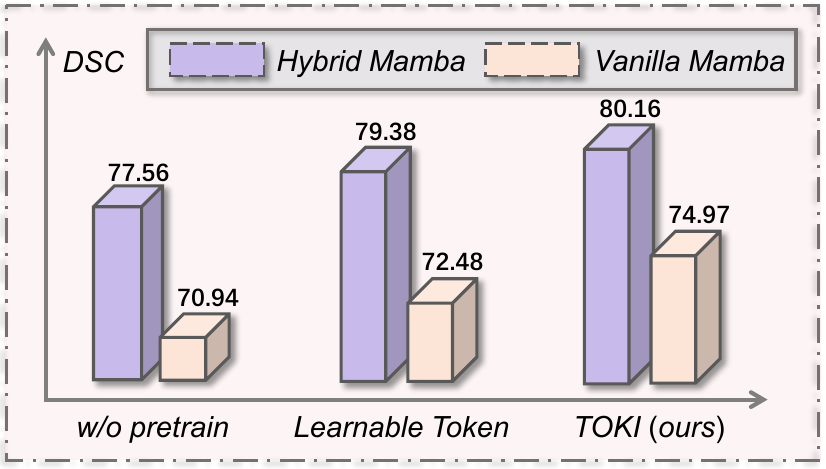}
    \caption{Different mask token strategies for Vanilla Mamba and Hybrid Mamba pre-trained on the BTCV dataset~\citep{btcv} for the 3D segmentation task. 
    The improvements brought by TOKI surpass previous SSL methods with the learnable token, and they are much better than those without pre-training.}
    \vspace{-4mm}
    \label{fig:example}
\end{figure}

\begin{figure*}
    \centering
    \includegraphics[width=0.85\linewidth]{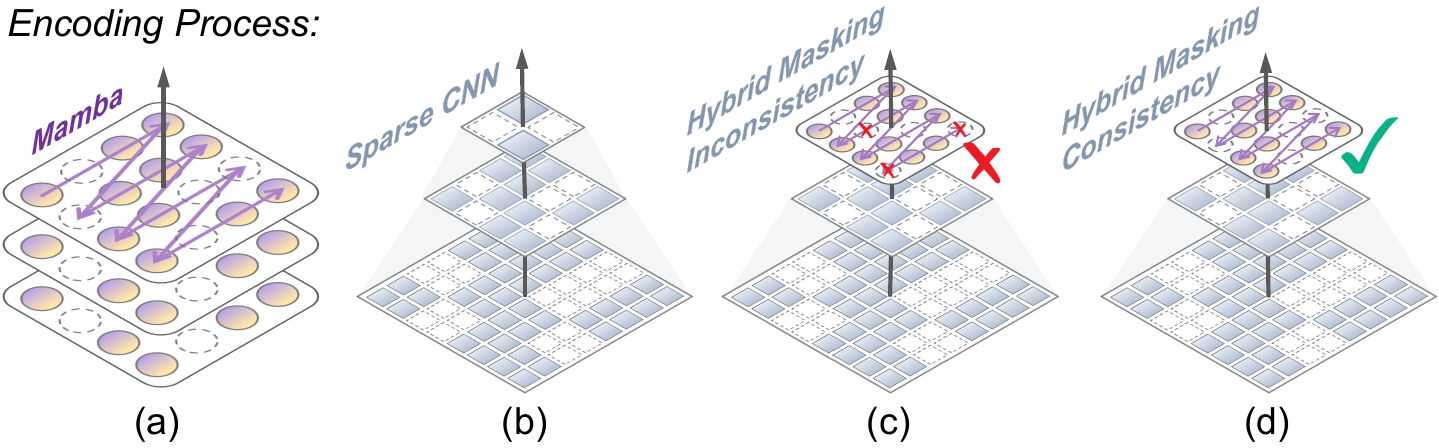}
    \vspace{-2mm}
    \caption{{The illustration of the masking inconsistency problem. (a) Directly dropping in Mamba. (b) Sparsely dropping in CNN. (c) Masking inconsistency in hybrid architecture. (d) Masking consistency in hybrid architecture. Mask inconsistency in hybrid architectures refers to the cross-architecture inconsistency in mask position that induces pixel intensity distributional shifts during encoding, which degrades representation learning.}} \label{fig:masking_incosistency}
    \vspace{-2mm}
\end{figure*}

The pre-training of Vision Mamba models using MIM remains an unrealized and largely unexplored area. Two primary challenges hinder applying masked image modeling on Vision Mamba:
\textbf{(1) Limited causal token learning ability in state space modeling}: As illustrated in Fig.\ref{fig:pipeline}(a), one straightforward idea for masked modeling Vision Mamba is to drop masked patches and insert learnable tokens at the masked positions to reconstruct the masked pixels, a technique previously utilized in MAE~\citep{mae} for Vision Transformers (ViTs)~\citep{vit}. However, our preliminary experiments indicate that while learnable tokens have effect on both vanilla and hybrid Mamba, their performance is not yet satisfactory (as shown in Fig.\ref{fig:example}, learnable token (pink)). We attribute this shortcoming to \textit{the insertion of learnable tokens that fail to comply with the causal and input-dependent selective scan property}. Since these tokens ignore structure sequence relationships within the state space, historical information is misrepresented by randomly initialized learnable tokens, which could fail to effectively convey information to subsequent tokens. \textbf{(2) Masking consistency across different architectures}:
{As shown in Fig.}\ref{fig:masking_incosistency} {, MIM requires maintaining positional consistency of mask regions throughout the encoder from input to output, where the pixel intensity distribution should remain invariant before and after masking~{\citep{spark}}. As for single architectures, different architectures employ specifically designed masking strategies: sparse convolution in CNNs skip-calculating mask positions (shown in Fig.} \ref{fig:masking_incosistency} {(b))~{\citep{spark, convnextv2}}, while Vision Transformers (ViTs) address the challenge by dropping or masking relation-independent tokens (shown in Fig.}\ref{fig:masking_incosistency}{ (a))~{\citep{mae, simmim}}. However, for hybrid architectures (including both serial and parallel combinations of different modules), due to their unique masking strategies, preserving mask position consistency across heterogeneous components presents a significant challenge. \textit{Inconsistent masking between architectural components could lead to shifts in pixel distribution} (shown in Fig.}\ref{fig:masking_incosistency}{ (c))~{\citep{spark}}, undermining the model's representation learning capability.}

To address the aforementioned challenges, we propose a Mamba-specific masking modeling strategy, MambaMIM, for pre-training single or hybrid Mamba-based models on large-scale 3D medical images. 
To deal with the limitation for casual learning, MambaMIM employs a distinctive token generation method called selective structure state space sequence {\bf TOK}en-{\bf I}nterpolation (TOKI) shown in Fig.\ref{fig:pipeline}(b). TOKI generates the mask token by effective utilization of relationships between the unmasked state space sequences. Instead of using randomly initialized tokens, we calculate the interpolation of adjacent masked tokens based on the state space context between unmasked tokens, where these interpolations are utilized as masked tokens for reconstruction. The computation of hidden state and output are \textbf{\textit{using the state space function}}, which encapsulate the relationships and dynamics of Mamba architecture. To keep masking consistency across different
architectures, MambaMIM introduces a novel  \textbf{\textit{bottom-up mask modeling strategy}}, which ensures that the masking positions are the same across different architectures and feature scales. Specifically, MambaMIM performs sparse operations for CNN and encodes visible sequence patches for Vision Mamba. Then, the hierarchical hybrid decoding is introduced to learn multi-scale representation. MambaMIM can be used to pre-train with any hybrid models and vanilla Vision Mamba models (a special case in hybrid models) which can be seen in Fig.~\ref{fig:example}.

Through the proposed MambaMIM, we successfully enable generative pre-training, benefiting both vanilla and hybrid Mamba architectures. To the best of our knowledge, MambaMIM is the first MIM method specifically designed for Mamba. Across multiple medical image segmentation tasks, the hybrid model, \textit{i.e.} a custom hybrid architecture where the top encoder is modern MedNeXt~\citep{mednext} and the bottom is Vision Mamaba, pre-trained with MambaMIM outperforms other state-of-the-art self-supervised pre-training methods and architectures. Our main contributions are:
\begin{itemize}
    \item 
    We propose MambaMIM, a generative self-supervised learning method for pre-training any single or hybrid Mamba architectures that can learn state-space representations. To the best of our knowledge, MambaMIM is the first attempt to explore the potential of MIM on Mamba.

    \item 
    We propose a novel type of token generation method called selective structure state space sequence token-interpolation (TOKI) for Mamba, which prioritizes consistency across the state space for causal learning and maximizes the potential of Mamba for long-range representation learning.

    \item
    We ensure mask consistency between CNN-Mamba architectures through bottom-up mask modeling and introduce hierarchical hybrid decoding to learn multi-scale representations.
    
    \item 
    We pre-train a strong hybrid vision encoder using MambaMIM on large-scale 3D CT datasets and fine-tune it on various downstream tasks. Extensive experiments demonstrate the effectiveness and potential of MambaMIM, achieving state-of-the-art performance.
\end{itemize}


\section{Related Works}
In this section, we first introduce the previous masked image modeling methods. Then, we survey the existing self-supervised methods in medical image analysis. Finally, we review the recent Mamba-based methods for medical tasks and discuss the importance of introducing a self-supervised method for Mamba.
\subsection{Masked image modeling}
Driven by BERT~\citep{beit}, masked image modeling (MIM) aims to remove or corrupt portions of the visual input and learn to predict the corrupted ones~\citep{inpaint, simmim, mae, spark, highlevel}. These approaches have been studied to reveal their ability to learn local attention patterns~\citep{lowlevel} and demonstrate better transferability to downstream tasks, such as segmentation and detection~\citep{mae, simmim, cmae, spark}. However, MIM methods necessitate handling both masked and seen patches, typically involving strategies tailored to specific architectures~\citep{simmim, mae, spark}. One representative MIM technique is MAE~\citep{mae}, which pre-trains Vision Transformer (ViT)~\citep{vit} by handling only a small subset of visible patches. This computationally efficient approach offers significant advantages for 3D medical SSL~\citep{survey, mae3d}. However, these methods ignore the structural sequence relationships within the state space, impairing Mamba's ability to learn effective representations.
Unlike previous works, MambaMIM emphasizes the importance of learning causal relationships within the state space, learning the state space relationships between unmasked tokens to generate masked tokens for pixel space reconstruction.
\subsection{Self-supervised learning for medical image analyses}
Due to the scarcity of labeled medical images, self-supervised learning (SSL) has emerged as a promising approach for medical image analysis~\citep{survey}. Existing medical SSL methods are primarily based on contrastive learning, typically employing strong data augmentation techniques, such as rotation~\citep{swinunetr, rotate} and multi-view crops~\citep{prlv2, unimiss, voco, gvsl, vox2vec, transvw}. However, most of these methods learn modality-related high-level semantic representations~\citep{highlevel}, which introduce significant biases in downstream tasks with different data distributions~\citep{sslmia, bias}. In contrast, introducing MIM methods for medical imaging pre-training~\citep{mim, mae3d} offers a promising avenue for addressing these challenges~\citep{survey, mim}. MIM can learn robust local patterns~\citep{local}, which yield substantial benefits for various downstream tasks~\citep{survey, mim}, such as medical image segmentation, while also offering improved generalization ability. However, there is currently no existing MIM approach specifically designed for Mamba, which is a promising architecture capable of efficiently handling long sequences of 3D medical images. 

\begin{figure*}[t]
    \centering
    \includegraphics[width = 0.99\linewidth]{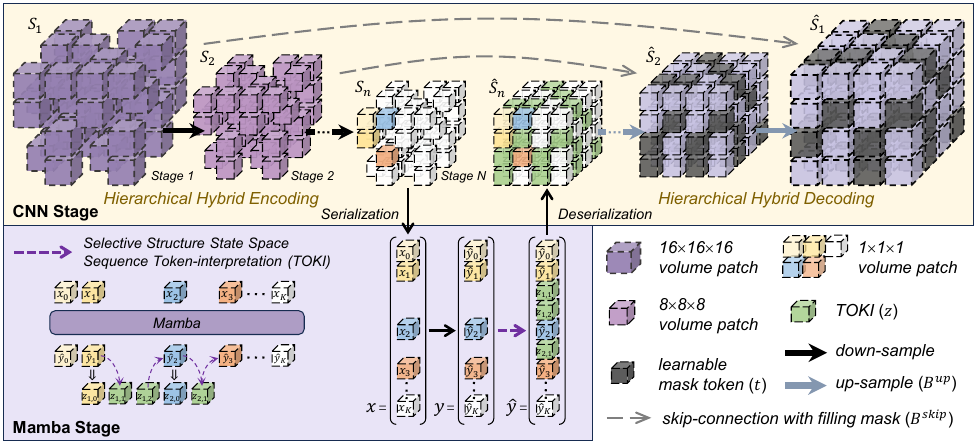}
    \caption{The whole structure of MambaMIM. The hybrid encoder performs bottom-up masked modeling, the initialization unmasking patch is white, the bottom-up mapping unmasking patch is purple and the masking position is empty. The CNN stage (yellow) utilizes 3D sparse operator for hierarchical encoding and fills learnable mask tokens in masked position for decoding. The Mamba stage (purple) only learns unmasked sequences, and TOKI is applied during decoding which can preserve the continuity of 1D selective structure state space sequence.}
    \label{fig:enter-label}
    \vspace{-2mm}
\end{figure*}

\subsection{Mamba in medical image analyses}
Long-range dependency modeling is critical for 3D medical image analysis tasks~\citep{vimsurvey, mambasurvey2}. In recent years, this capability has been enabled by Transformers~\citep{transformer}. However, Transformers exhibit quadratic complexity with respect to sequence length, which can pose a computational challenge for high-resolution 3D medical images~\citep{vimsurvey, mambasurvey2}. Recently, Mamba, with its hybrid architectures, has been introduced for medical image analysis tasks, such as classification~\citep{cls}, reconstruction~\citep{reconstruction}, registration~\citep{registration}, and segmentation~\citep{umamba, segmamba, swinumamba}, demonstrating comparable performance to Transformers while significantly reducing computational overhead. Given the labor-intensive annotations of medical images~\citep{abd1k, word} and the importance of efficient long-range dependency modeling in 3D medical image analysis~\citep{vimsurvey, mambasurvey2}, pre-training Mamba and transferring it for downstream tasks presents a promising solution for label-efficient learning.

\section{Methodology}
To solve the problem of pre-training methods for any hybrid mamba, we proposed a  generative self-supervised learning method MambaMIM. The overall framework of MambaMIM is illustrated in Fig.~\ref{fig:enter-label}, which consists of two components: hierarchical hybrid encoding (Sec~\ref{sec:encoding}) and decoding (Sec~\ref{sec:decoding}) are used for bottom-up hybrid mask modeling; Selective Structured State Space Sequence Token-interpolation (Sec~\ref{sec:TOKI}) is used for learning causal relationships within state space modeling. Aside from the Vision Mamba baseline, we further propose to use a CNN-Mamba hybrid model, called HyMamba (CNN on the top and Vision Mamba at the bottom) as the baseline backbone, to fully reveal the potential of our proposed MambaMIM and Mamba structure.

\subsection{Baseline models}

\noindent{\textbf{Vanilla Mamba.} For vanilla Mamba, we utilize the UNETR~{\citep{unetr}} network, where the Vision Transformer (ViT) is replaced by Vision Mamba~{\citep{vim}}. Following the patch embedding practice in UNETR, we set the patch size of $16\times16\times16$, depth of $12$ and embedding dims of $768$, and use the same bottleneck and decoder of UNETR in vanilla Mamba.}

\noindent{\textbf{HyMamba.} For hybrid Mamba (HyMamba), the encoder adopts a serial architecture consisting of two stages: the CNN stage and the Mamba stage. In the CNN stage, we choose MedNeXt~{\citep{mednext}} (the state-of-the-art ConvNet in medical tasks) as the top part of the encoder. In the Mamba stage, we implement the standard Vision Mamba~{\citep{vim}} as the bottom part of the encoder. It is worth noting that we substitute the downsampling layers of MedNeXt with max-pooling. For downstream tasks, we utilize the MedNeXt's decoder for segmentation. we set hierarchical CNN with 4 stages and the patch size is 16 $\times$ 16 $\times$ 16 during pre-training, fine-tuning, and testing.}

\subsection{Hierarchical hybrid encoding}
\label{sec:encoding}
In developing an end-to-end pre-training CNN-Mamba hybrid model, a key challenge is ensuring consistent masking across the CNN and Mamba stages of the encoder. To address this, we propose a bottom-up hybrid masking strategy. Specifically, as shown in Algorithm \ref{alg:algorithm}, given the input volume $X \in \mathbb{R}^{H \times W\/  \times D}$ and {a hierarchical CNN with {$n$} stages (stage {\textit{1}}-th to stage {\textit{n}}-th)}, we initialize the mask $M_n \in \mathbb{R}^{\frac{H}{n^2} \times \frac{W}{n^2}\/  \times \frac{D}{n^2}}$ at the $n$-th CNN stage {(the final CNN stage)}. Then, we map the initialized mask $M_n$ to $\{1,2, ..., n-1\}$-th stages and generate a set of masks $\{ M_1, M_2, \cdots, M_{n-1}  \}$ by up-sampling masked patches at different scales in the same manner.
For the CNN stage, following previous work~\citep{spark,convnextv2}, we introduce 3D sparse operator (SparseOp) to calcuate the masked positions and get different scale sparse features $\{ S_1, S_2, \cdots, S_n \}$. {The definition of the sparse operator is as follows:}

\begin{equation}
    \text{SparseOp} (S_{i-1},M_{i}) = \text{Operator} (S_{i-1}) \cdot M_{i}
\end{equation}

\noindent {where $S_{i-1}$ is the $(i-1)$-th CNN stage sparse feature and $M_{i}$ is the $i$-th stage mask. The operator refers to the spatial operations (\textit{i.e.}, convolution~{\citep{spconv}} and normalization~{\citep{norm}}) on sparse features.} In the Mamba stage, the sparse feature $S_n$ is serialized into a sequence and the unmasked tokens $x$ of $S_n$ are obtained. Then the unmasked tokens $x$ are used to generate the state-space sequence $y$. Notably, the final output of the encoder consists of hierarchical sparse features $\{ S_1, S_2, \cdots, S_{n-1}, y\}$ from both CNN and Mamba, following the same masking strategy. The detailed encoding process is shown in Algorithm \ref{alg:algorithm}.

\begin{algorithm}[t!]
\caption{Hierarchical hybrid encoding}
\label{alg:algorithm}
\textbf{Input}: Image volume $X$\\
\textbf{Output}: A set of sparse features $\{y, S_{n-1}, ..., S_2, S_1\}$\\
\textbf{Initialize}: Random masking $M_n$
\begin{algorithmic}[1] 
\STATE $\{ M_{n-1}, M_{n-2}, ..., M_{2}, M_{1}\} \gets M_n$
\STATE $S_1 \gets \text{SparseOp}(X, M_1)$
\STATE \textbf{for} $i = 2$ \textbf{to} $n$ \textbf{do}
\STATE \quad \quad  $S_i \gets \text{SparseOp}(S_{i-1}, M_i)$
\STATE \textbf{end for}
\STATE $x \gets \text{Serialization}(S_n, M_n)$
\STATE $y \gets \text{Mamba}(x)$
\RETURN $\{y, S_{n-1}, ..., S_2, S_1\}$
\end{algorithmic}
\end{algorithm}

\subsection{Hierarchical hybrid decoding} 
\label{sec:decoding}
To effectively utilize this feature of Mamba, it is necessary to implement different mask-filling strategies according to the characteristics of each architecture. For Mamba, a significant property can be derived from the SSM formula~\citep{gu2021efficiently}, demonstrating that adjacent blocks exhibit interactions within the state space, a characteristic not present in CNN. Specifically, for the Mamba stage, a novel state-space interpolation TOKI is proposed to preserve the continuity of 1D selective structure state space sequences (in Sec.~\ref{sec:TOKI}). As for the CNN stage, we introduce a simple cascaded decoder comprising $N$-$1$ upsampling blocks $\{B_{1}^{up}, B_{2}^{up},...,B_{n-1}^{up}\}$ and skip-connection blocks $\{B_{1}^{skip}, B_{2}^{skip},...,B_{n-1}^{skip}\}$ for hierarchical dense reconstruction. {Specifically, the upsampling block consists of two convolutional layers and an upsampling layer, while the fusion block comprises two convolutional layers. During hierarchical dense reconstruction,} we fill learnable token embeddings $\{t_{n-1}, ..., t_2, t_1\}$ into all empty positions of 3D sparse features $\{S_1, S_2, \cdots, {S}_{n-1} \}$ at different scales to get dense features  $\{\hat{S_1}, \hat{S_2}, \cdots, \hat{S}_{n-1} \}$. 
Finally, we utilize a linear layer to reconstruct results $\hat{X}$. 
The detailed decoding process is presented in Algorithm \ref{alg:decoding}.

\begin{algorithm}[t!]
\caption{Hierarchical hybrid decoding}
\label{alg:decoding}
\textbf{Input}: A set of sparse features $\{y, S_{n-1}, ..., S_2, S_1\}$\\
\textbf{Output}: Reconstruction map $\hat{X}$ \\
\textbf{Operator}: {projection {$\phi$}}, linear layer $\tau$ \
\begin{algorithmic}[1] 
\STATE $\hat{y} \gets \text{TOKI}(y)$
\STATE $D_n \gets \phi(\hat{y})$ 
\STATE \textbf{for} $i = n-1$ \textbf{to} $1$ \textbf{do}
\STATE \quad \quad $\hat{S_{i}} \gets \text{Fill}(S_{i}, t_{i})$ 
\STATE \quad \quad $D_i \gets B_{i}^{skip}(Concat\{\phi(\hat{S_{i}}), B_{i}^{up}(D_{i+1})\}) $
\STATE \textbf{end for}
\STATE $\hat{X} \gets \tau(D_1)$
\RETURN $\hat{X}$
\end{algorithmic}
\end{algorithm}

\subsection{Selective structured state space sequence token-interpolation (TOKI)}
\label{sec:TOKI}

{Mamba relies on a state space model (SSM), where each output depends on previous states in a continuous, structured manner. However, traditional methods such as MAE~{\citep{mae}} face challenges when applied to Mamba, as they often fill masked regions with random or learnable tokens, which disrupts the causal flow of a sequence, making it harder for state-space models to learn high-quality representations. TOKI addresses this key challenge in applying MIM to Mamba-based networks. Unlike traditional methods that insert learnable tokens into masked regions, TOKI leverages the inherent causal relationships within Mamba's state-space sequences and mathematically interpolates the masking state-space token, which ensures the filled tokens respect the causal flow of the sequence.}

\begin{algorithm}[tb]
\caption{Selective structure state space sequence \\ token-interpretation (TOKI)}
\label{alg:TOKI}
\textbf{Input}: $y = [\hat{y_{i}},\hat{y}_{i+1}], Q+1, $\\
\textbf{Parameter}: Learnable parameters $\bar{A'}$ \\
\textbf{Operator}: Exponentiation $\text{Pow}(\cdot)$ \\
\textbf{Output}: TOKI sequence $\hat{y}$
\begin{algorithmic}[1] 
\STATE \textbf{for} $j = 1$ \textbf{to} $Q+1$ \textbf{do}
\STATE \quad \quad $\alpha \gets \frac{j}{Q+2}$
\STATE \quad \quad $V \gets (1-\alpha) \cdot \hat{y_{i}} + \alpha \cdot \hat{y_{i+1}}$
\STATE \quad \quad $z_{j} \gets \sum_{n=0}^{j} \text{Pow}(A',j-n) \Delta A'^{-1} \cdot V$

\RETURN $\hat{y} \gets [\hat{y_{i}} = z_{0}, z_{1}, z_{2},..., z_{Q+1}]$
\end{algorithmic}
\end{algorithm}


\noindent\textbf{Problem definition and analysis.} In modern state-space models, structured state-space sequence models (S4) and Mamba both rely on a classical continuous system that maps one-dimensional input function or sequence, denoted as $x(t) \in \mathcal{R}$, through inter-mediate implicit state $h(t) \in \mathcal{R}^{N}$ to an output $y(t) \in \mathcal{R}$. The whole process can be represented as a linear Ordinary Differential Equation (ODE):
\begin{align}\begin{aligned}
\label{Equ.originFormula}
    h'(t) &= Ah(t)+Bx(t), \\
    y(t) & = Ch(t),
\end{aligned}\end{align}
where $A \in \mathcal{R}^{N \times N}$ means the state matrix like combination coefficient, $B \in \mathcal{R}^{N \times 1}$ and $C \in \mathcal{R}^{N \times 1}$ represents computation parameters. To make it more suitable for deep learning scenarios, discretization is performed. A timescale parameter $\Delta$ is utilized. And $\bar{A}$ and $\bar{B}$ with zero-order hold can be defined as follows:
\begin{align}\begin{aligned}
    \bar{A} &= exp(\Delta A), \\
    \bar{B} & = (\Delta A)^{-1}(exp(\Delta A) - I) \cdot \Delta B, \\
    \Delta & = Linear(x),
\end{aligned}\end{align}
The time-scale parameter $\Delta$ is derived from $x$, creating a data dependency on $\Delta$. Consequently, this introduces time-varying characteristics to the entire process.

After discretization, SSM-based models can be computed as following way: global convolution, defined as Equ.~\ref{Equ:MartixResp}:
\begin{align}\begin{aligned}
\label{Equ:MartixResp}
    \bar{K} &= (C\bar{B}, C\bar{A}\bar{B},.....,C\bar{A}^{L-1}\bar{B}), \\
    y &= x * \bar{K},
\end{aligned}\end{align}
where $\bar{K} \in R^{L}$ represents a structure convolutional kernel, and $L$ is the length of the input sequence $x$.

To facilitate theoretical deduction, we express Equ.~\ref{Equ:MartixResp} as a series of summations.
\begin{align}\begin{aligned}
\label{Equ.Summations}
    y &= (y_{1},y_{2},..., y_{j},...,y_{L}), \\
    y_{j} &= \sum_{i=1}^{j} C \cdot \bar{A} ^{j-i} \cdot \bar{B} x_{i}. \\
\end{aligned}\end{align}
Taking masked image modeling into consideration, $\Omega$ represents the masked sequence index. Define $\Gamma_{\Omega}(i)$ as the querying the index of the $i$-th elemnet of $y$ in the sequence ${\Omega}$ and Equ.~\ref{Equ.Summations} can be further expressed via:
\begin{align}\begin{aligned}
\label{Equ.y}
    \hat{y} & = ( \hat{y}_{1}, \hat{y}_{2},.....,\hat{y}_{L}), \\
    \hat{y_{i}} &= \begin{cases}
                \sum_{n=1}^{j} C \cdot \bar{A}^{j-n} \cdot \bar{B} x_{j}, j =\Gamma_{\Omega}(i),\text{if } i \in { \Omega}  \\
                \delta,\text{if } i \in \bar{\Omega}
                  \end{cases},
\end{aligned}\end{align}
where $\delta$ represents learnable parameters, which are often set to identity matrix. It can be seen that $\delta$ helps the encoder to learn sparse semantic information, but it has a possibility of failure for the intrinsic relationships of the structure state space. The success of Mamba on vision tasks relies on exploration of inter-mediate implicit states. Simply filling masked sequence with tokens $\delta$ may not satisfy this mathematical relation and is not suitable for Mamba. 
\textit{In order to enable filled sequences to meet with causal relationships in structure state space, TOKI is proposed.} {TOKI identifies the visible tokens before and after the masked region and mathematically interpolates between these visible states, ensuring that the filled state-space tokens follow} {the same causal flow as the rest of the sequence. This preserves Mamba’s ability to model long-range dependencies efficiently, enabling better representation learning during pre-training.}

\noindent\textbf{TOKI derivation.}
We interpolate sequences and they can be rearranged into the following form. Given the masked input sequence $x = (x_{1},...,x_{K})$, masked Mamba sequence $y = (\hat{y_{1}},...,\hat{y_{K}})$ is achieved after Mamba block. We present a generalizable masked decoding scenario:
\begin{align}\begin{aligned}
\label{Equ.AnlysisFormula}
    \hat{y} &= (\hat{y}_{1},...,\hat{y}_{i},z_{1},z_{2},...,z_{Q},\hat{y}_{i+1},\hat{y}_{i+2},....,\hat{y}_{K}), \\
    T & = Q+K, \\
    \hat{y_{j}} &= \sum_{n=1}^{j} C \cdot \bar{A}^{j-n} \cdot \bar{B} s_{n} ,\\
    z_{i} &= \delta, 
\end{aligned}\end{align}
where $\hat{y}$ is the decoding sequence, $K$ is the number of the unmasked tokens in Mamba sequence, $Q$ is the number of the filled tokens and $T$ is the length of the whole reconstruction sequences.
We hypothesize that if the position of $z_{i}$ is left unfilled by a token, the entire sequence forms a complete selective state space. To restore this deductive relationship, we assume that the sequence $(\hat{y}_{i} = z_{0}, z_{1}, z_{2}, ..., z_{Q}, \hat{y}_{i+1}= z_{Q+1})$ also satisfies this relationship. By adding the terms $\hat{y}_{i}$ at the beginning and $\hat{y}_{i+1}$ at the end, the index of the sequences is extended to \([0, Q+1]\), thereby increasing its total length to \(Q+2\).  The mathematical expression of selective state space token can be represented as:

\begin{align}\begin{aligned}
\label{Equ.RewriteFormula}
    \hat{y_{i}} = z_{0} &= C' \bar{B'} s'_{0},\\
    z_{i} & = \sum_{n=0}^{i} C' \cdot \bar{A'}^{i-n} \cdot \bar{B} s'_{n}, \\
    \hat{y}_{i+1} = z_{Q+1} & = \sum_{n=0}^{Q+2} C' \cdot \bar{A'}^{Q+2-n} \cdot \bar{B'} s'_{n}, \\
    s'_n &= \frac{Q+2-n}{Q+2} \cdot \hat{y}_{i} + \frac{n}{Q+2} \cdot \hat{y}_{i+1},
\end{aligned}\end{align}
where $C', \bar{B}', \bar{A}'$ can be set as learnable parameters, and theoretically, all the values are solvable. However, obtaining these values is challenging for the network. We further observe that if we set $A'$ as learnable parameters, the effects of $C'$ and $B'$ could be represented by $A'$ via gradient update. For simplification, let $B' = C' = E$ (identity matrix).  
Then $\bar{B}'$ can be written as $(e^{\Delta A'} \Delta A'^{-1} - \Delta A'^{-1}) \approx e^{\Delta A'} \Delta A'^{-1}$.

\section{Experiment}
\subsection{Dataset}

\noindent\textbf{Pre-training datasets.} As shown in Table.\ref{tab:dataset}, our pre-training dataset includes 6814 CT scans of various body region of interest (ROI) such as head, neck, chest, abdomen, and pelvis. It is composed of 13 public datasets: BTCV~\citep{btcv}, CHAOS~\citep{chaos}, WORD~\citep{word}, FLARE'22~\citep{flare22}, AbdomenCT-1k~\citep{abd1k}, TotalSegmentator~\citep{totalseg}, AMOS~\citep{amos}, and MSD~\citep{msd}. Existing labels are not utilized from these datasets during the pre-training stage.

\noindent\textbf{BTCV dataset.} The Multi-Atlas Labeling Beyond The Cranial Vault (BTCV)~\citep{btcv} dataset consists of 30 subjects with abdominal CT scans where 13 organs are annotated at pixel level by interpreters under the supervision of clinical radiologists at Vanderbilt University Medical Center. Our data preprocessing strategy is the same as UNETR~\citep{unetr}.

\noindent\textbf{MSD dataset.} Medical Segmentation Decathlon (MSD) dataset~\citep{msd} comprises of ten segmentation tasks from different organs and image modalities. We only
use three CT datasets: Pancreas, Hepatic Vessel, Spleen dataset. All the pre-processing strategy is the same as Swin UNETR~\citep{swinunetr}.

\noindent\textbf{AMOS dataset.} The Multi-Modality Abdominal Multi-Organ Segmentation Challenge (AMOS) dataset~\citep{amos} comprises 360 CT and MRI scans annotated for 15 abdominal organs. The data preprocessing strategy is the same as Swin UNETR~\citep{unetr}.

\noindent {\textbf{CT-ORG dataset.} The CT-ORG dataset~{\citep{ctorg}} consists of 150 CT scans for the five organ and bone annotations from several clinical sites. It is used for multiple organ segmentation.}

\noindent {\textbf{KiTS 23 dataset.} The 2023 Kidney and Kidney Tumor Segmentation challenge (KiTS 23) dataset~{\citep{kits21}} consists of 489 CT scans for the Kidney and its Tumor and Cyst annotations from M Health Fairview medical center.}

\noindent\textbf{BraTS 21 dataset.} The BraTS 21 dataset~\citep{brats21}, from the BraTS 2021 challenge of brain tumors by providing 1251 MRI scans with pixel-level annotations. Following the previous works~\citep{swinunetr, voco}, we utilize the same dataset split, \textit{i.e.}, with 1000 scans for training and 251 scans for validation, for fair comparison. The data preprocessing strategy is the same as Swin UNETR~\citep{unetr}.

\begin{table}[!t]
\caption{Overview of pre-train dataset.\label{tab:dataset}}
\vspace{-2mm}
\centering
\resizebox{0.95\linewidth}{!}
{
\begin{tabular}{l c c c}
\Xhline{1px} 
Dataset &  \# of volumes\\
\hline
BTCV~\citep{btcv} & 50  \\
CHAOS~\citep{chaos} & 40  \\ 
WORD~\citep{word} & 150  \\
FLARE'22~\citep{flare22} & 2300  \\
AbdomenCT-1k~\citep{abd1k} & 1062  \\
TotalSegmentator \citep{totalseg} & 1202 \\
MSD Spleen~\citep{msd} & 61  \\
MSD Lung~\citep{msd} & 95  \\
MSD Pancreas~\citep{msd} & 420  \\
MSD Hepatic Vessel~\citep{msd} & 443  \\
MSD Liver~\citep{msd} & 201  \\
MSD Colon~\citep{msd} & 190  \\
\hline
Total & 6814   \\
\Xhline{1px}
\end{tabular}
}
\vspace{-4mm}
\end{table}

\subsection{Implementation details}

\noindent\textbf{Implementation of pre-training.} The pre-training data are interpolated to isotropic voxel spacing of 1.5 mm. Intensities are scaled to [-175, 250] and then normalized to [0,1]. We crop sub-volumes of 96 $\times$ 96 $\times$ 96 voxels. For pre-training tasks, we train with the AdamW optimizer, a learning rate of $1e$-$4$, and a cosine-annealing learning rate scheduler. The batch size is set to 8, and the model is trained for 100 epochs on a single GPU. 

\noindent {\textbf{Implementation of fine-tuning.} For downstream segmentation tasks, a five-fold cross-validation strategy is used to train models for BTCV and MSD experiments, and the best model is selected on each fold. Additionally, for AMOS, we adhere to the official split, using 240 samples for training and 120 samples for validation. For CT-ORG and KiTS 23, we adopt an 80/20 train-validation split. To assess the model's cross-modality generalization capability, we transfer the pre-trained model from the CT domain to the MRI (i.e. adapt in BraTS 21~{\citep{brats21}}) for further evaluation. We random crop sub-volumes of $96 \times 96 \times 96$ voxels by ratios of positive and negative as 1:1 in 4 sub-crops. Data augmentations are the same as Swin UNETR~{\citep{swinunetr}}. For the segmentation task, we initialize the encoder with pre-trained weights and fine-tune the entire network using the AdamW optimizer with a learning rate of $1e$-$4$ and fine-tuning for 50,000 iterations. All other fine-tuning details are consistent with Swin UNETR~{\citep{swinunetr}}. All methods are implemented in PyTorch and trained on a Nvidia A800 GPU.}

\noindent\textbf{Loss function.}  During the pre-training, given the output prediction of reconstruction $\hat{X}$ and input volume $X$, a mean square error loss ($\mathcal{L}_{2}$) is for the masked positions reconstruction. During fine-tuning, we use a combined loss ($\mathcal{L}_{seg} = \mathcal{L}_{BCE} + \mathcal{L}_{DSC}$) of binary cross entropy loss ($\mathcal{L}_{BCE}$) and DSC loss ($\mathcal{L}_{DSC}$) to optimize the network.

\renewcommand{\multirowsetup}{\centering}  
\begin{table*}[h!]
\centering
\caption{Results of the baseline backbone on BTCV. \textbf{val} (bold) / \underline{val} (underline) : top method / second method\label{tab:btcv}.}
\vspace{-2mm}
\resizebox{1\linewidth}{!}
{
\begin{tabular}{l | c | ccccccccccc| c}
\Xhline{1px}
\multicolumn{1}{c |}{Network} & \multicolumn{1}{c |}{Pre-training}  & \multirow{1}{*}{Spl} & \multirow{1}{*}{Kid} & \multirow{1}{*}{Gall} & \multirow{1}{*}{Eso} & \multirow{1}{*}{Liv} & \multirow{1}{*}{Sto} & \multirow{1}{*}{Aor} & \multirow{1}{*}{IVC} & \multirow{1}{*}{Veins} & \multirow{1}{*}{Pan} & \multirow{1}{*}{AG} & \multirow{1}{*}{Avg (\%)} \\
\cline{1-2}
\hline
UNETR &\ding{55} & 86.14 & 82.73 & 55.99 & 67.23 & 92.50 & 73.80 & 85.26 & 77.59 & 59.81 & 53.71 & 53.75 & 71.15 \\ 
Swin UNETR &\ding{55} & 88.51 & 81.85 & 58.39 & 71.72 & 94.18 & 78.47 & 86.97 & 80.51 & 66.96 & 64.71 & 60.13 & 74.95 \\ 
MedNeXt &\ding{55}  & 89.27 & 83.88 & 60.85 & 71.66 & 94.30 & 79.39 & \underline{88.94} & 82.25 & 64.85 & 70.63 & 59.94 & 76.14 \\
SegMamba &\ding{55}   & 88.19 & 82.11 & 61.52 & 70.54 & 93.29 & 80.13 & 87.24 & 82.81 & 66.31 & \underline{72.38} & \underline{64.23} & 76.54 \\
3D U-Net (FPN) &\ding{55}  & 88.77 & 80.54 & 61.76 & \underline{74.40} & \underline{94.88} & 80.93 & 88.25 & \underline{83.21} & 67.68 & 71.92 & 62.22 & 76.72 \\ 

    {LightM-UNet} & \ding{55} & {82.04} & {75.55} & {55.44} & {65.32} & {91.48} & {64.97} & {80.32} & {70.95} & {60.22} & {59.06} & {53.67} & {68.33} \\

{MedSegMamba} & \ding{55} & {\textbf{89.86}} & {84.17} & {58.57} & {74.30} & {94.03} & \underline{{83.47}} & {88.53} & {81.76} & {65.99} & {69.11} & {61.72} & {76.72} \\

\hline \multirow{2}{*}{Vanilla Mamba} &\ding{55} & 85.08 & 82.61 & 58.65 & 68.25 & 92.66 & 72.35 & 84.12 &75.94 & 56.68 &53.34 & 52.96 & 70.94 
\\
 & \checkmark(MambaMIM) & 86.78 & 84.85
 & \underline{62.68} & 69.58 & 92.94 & 79.45 & 87.66 & 79.99 & 61.88 & 65.30 &  59.40 & 74.97 \\
\hline
 \multirow{2}{*}{HyMamba} &\ding{55}  & 88.60 & \underline{87.18} & 60.67 & \textbf{74.73} & 94.34 & 79.60 & 88.67 & 82.07 & \underline{68.83} & 71.53 & 62.43 & \underline{77.56} \\ 
 & \checkmark(MambaMIM)  & \underline{89.67} & \textbf{87.23} & \textbf{65.93} & 73.91 & \textbf{95.00} & \textbf{86.20} & \textbf{91.08} & \textbf{84.26} & \textbf{70.40} & \textbf{79.11} & \textbf{64.81} & \textbf{80.16} \\ 
\Xhline{1px}
\end{tabular}
}
\end{table*}

\renewcommand{\multirowsetup}{\centering}  
\begin{table*}[h!]
\centering
\caption{Comparison between MambaMIM and other pre-training methods with the HyMamba on BTCV.  \textbf{val} (bold) / \underline{val} (underline) : top method / second method.\label{tab:btcv3}}
\vspace{-2mm}
\resizebox{1\linewidth}{!}
{
\begin{tabular}{l | c | ccccccccccc| c}
\Xhline{1px} 
\multicolumn{2}{c |}{Pre-training method}  & \multirow{2}{*}{Spl} & \multirow{2}{*}{Kid} & \multirow{2}{*}{Gall} & \multirow{2}{*}{Eso} & \multirow{2}{*}{Liv} & \multirow{2}{*}{Sto} & \multirow{2}{*}{Aor} & \multirow{2}{*}{IVC} & \multirow{2}{*}{Veins} & \multirow{2}{*}{Pan} & \multirow{2}{*}{AG} & \multirow{2}{*}{Avg (\%)} \\
\cline{1-2}
\multicolumn{1}{c| }{Method} & \multicolumn{1}{c| }{Backbone} &&&&&&&&&&&& \\

\hline

$\text{MoCov2}$ & \multirow{1}{*}{HyMamba} & 88.72 & 82.63 & 65.00 & 73.26 & 93.48 & 76.09 & 85.53 & 79.92 & 66.93 & 63.09 & 59.55 & 75.11 \\ 

$\text{SimSiam}$ & \multirow{1}{*}{HyMamba} & 88.50 & 85.46 & 65.31 & \textbf{75.22} & 94.15 & 77.35 & 88.66 & 82.12 & 68.88 & 71.65 & 62.20 & 77.48 \\ 

$\text{SUP}$ & \multirow{1}{*}{HyMamba} & 88.52  &84.06 & \textbf{68.45} &73.36 &93.82 & 80.78 & \underline{89.52} &81.89 &66.81 &69.54 & \underline{63.42} &77.83 \\

$\text{SwAV}$ & \multirow{1}{*}{HyMamba} & \textbf{89.78} & \textbf{88.35} & 65.38 & \underline{74.40} & \underline{94.54} & 79.29 & 88.42 & 82.23 & \underline{69.96} & \underline{73.11} & 61.91 & \underline{78.28} \\ 

{VoCo} & \multirow{1}{*}{{HyMamba}} & {88.76} & {84.52} & {64.75} & {72.63} & {94.48} & \underline{{81.08}} & {89.35} & \underline{{82.95}} & {66.75} & {68.24} & {62.51} & {77.16} \\

\rowcolor{gray!15} $\text{MambaMIM}$  & \multirow{1}{*}{HyMamba} & \underline{89.67} & \underline{87.23} & \underline{65.93} & 73.91 & \textbf{95.00} & \textbf{86.20} & \textbf{91.08} & \textbf{84.26} & \textbf{70.40} & \textbf{79.11} & \textbf{64.81} & \textbf{80.16}  \\
\Xhline{1px}
\end{tabular}
}

\end{table*}

\renewcommand{\multirowsetup}{\centering}  
\begin{table*}[!t]
\centering

\caption{Best performance of other state-of-the-art pre-training methods with network on BTCV and MSD. \textbf{val} (bold) / \underline{val} (underline) : top method / second method. $\dagger$ denotes we utilize official pre-training weights.\label{tab:msd}}
\vspace{-2mm}
\resizebox{1\linewidth}{!}
{
\begin{tabular}{l | l | ccc | ccc | c | cccccc | c}
\Xhline{1px} 

\multicolumn{2}{c|}{Pre-training method} 
& \multicolumn{3}{c|}{ MSD Pancreas}
& \multicolumn{3}{c|}{ MSD Hepatic Vessel}
& \multicolumn{1}{c|}{MSD Spleen}
& \multicolumn{6}{c|}{BTCV}
&  \multirow{2}{*}{Avg (\%)}
\\
\cline{1-15}
\multicolumn{1}{c|}{Method} & \multicolumn{1}{c|}{Network}  & \multicolumn{1}{c}{DSC1} & \multicolumn{1}{c}{DSC2} & \multicolumn{1}{c|}{Avg (\%)} & \multicolumn{1}{c}{DSC1} & \multicolumn{1}{c}{DSC2} & \multicolumn{1}{c|}{Avg (\%)} & \multicolumn{1}{c|}{DSC} & \multirow{1}{*}{Gall} & \multirow{1}{*}{Sto} & \multirow{1}{*}{IVC} & \multirow{1}{*}{Pan} & \multirow{1}{*}{AG} & \multirow{1}{*}{Avg (\%)} & \\
\hline
$\text{vox2vec}^{\dagger}$           & 3D U-Net             & 77.00 & 31.80 & 54.40 & 59.50 & 62.40 & 60.95 & 96.10 & 59.50 & 83.20 & \underline{83.90} & 73.90 & \textbf{65.20} & \underline{79.50}  & 72.73\\
$\text{SUP}^{\dagger}$ & Swin UNETR & 75.20 & 35.90 & 55.55 & 60.90 & 57.50 & 59.20 & 95.50 &  58.40 & 76.00 & 82.10 & 69.80 & 61.00 & 75.80  & 71.51 \\
$\text{MAE}$      & UNETR          & 77.76 & 39.29 & 58.52  & 59.99 & 62.22 & 61.10  & 95.28 & \underline{62.50} & \underline{86.11} & 83.26 & 75.47 & 63.77 & 79.07  & 73.49 \\
$\text{SimMIM}$    & Swin UNETR     & 76.16 & 44.96 & 60.56 & 60.67 & 61.79 & 61.23 & 95.64 & 60.28 & 78.42 & 81.46 & 66.34 & 58.65 & 74.73 & 73.04 \\
$\text{SparK}$  & MedNeXt      & \underline{78.88} & \underline{47.86} & \underline{63.37}  & \underline{61.08} & \underline{67.76} & \underline{64.42} & \underline{96.18} &  62.48 & 84.85 & 83.60 & \underline{76.57} & 64.13 & 79.21  & \underline{75.79} \\
\rowcolor{gray!15} $\text{MambaMIM}$ & HyMamba  & \textbf{78.90} & \textbf{51.01} & \textbf{64.96} & \textbf{61.61} & \textbf{68.22} & \textbf{64.91} & \textbf{96.41} &  \textbf{65.93} & \textbf{86.20} & \textbf{84.26} & \textbf{79.11} & \underline{64.81} & \textbf{80.16} &  \textbf{76.61} \\
\Xhline{1px}
\end{tabular}
}
\vspace{-4mm}
\end{table*}

\noindent\textbf{Evaluation metrics and comparison methods.} The Dice similarity coefficient (DSC) metric is used for all experimental results~\citep{lay2013rapid}. For comparison methods, we select three advanced general MIM strategies: Transformer-based MAE~\citep{mae} and SimMIM~\citep{simmim}, CNN-based SparK~\citep{spark}, three well-known general contrastive-based SSL methods: SimSiam~\citep{simsiam}, SwAV~\citep{swav}, MoCoV2\citep{moco}, and eight medical SSL method: Models Genesis (MG)~\citep{mg}, TransVW~\citep{transvw}, UniMiSS~\citep{unimiss}, \textbf{S}win \textbf{U}NETR \textbf{P}retrain method (SUP)~\citep{swinunetr}, PCRLv2~\citep{prlv2}, GVSL~\citep{gvsl}, vox2vec~\citep{vox2vec} and VoCo~\citep{voco}. In addition, we choose the current well-known segmentation networks UNETR~\citep{unetr}, Swin UNETR~\citep{swinunetr}, MedNeXt~\citep{mednext}, 3D U-Net~\citep{unet}, MiT~\citep{unimiss}, LightM-UNet~\citep{lightmunet}, MedSegMamba~\citep{medsegmamba}, Vanilla Mamba~\citep{mamba} and SegMamba~\citep{segmamba} for segmentation.

\subsection{Results and discussion}

\noindent\textbf{Performance of MambaMIM.} 
We first validate the proposed method MambaMIM and its baseline model on BTCV. The results are listed in Table~\ref{tab:btcv}. Our HyMamba achieves the highest average metric of 77.56\% when trained from scratch, surpassing Mamba by 6.62\% (77.56\% vs. 70.94\%) and SegMamba by 1.02\% (77.56\% vs. 76.54\%). This shows that the hybrid architecture can maximize the long-range extraction capability of Mamba, which is crucial for medical image segmentation tasks. More importantly, MambaMIM gains a huge leap in DSC score (2.6\% improvement, 80.16\% vs 77.56\%). For organs that are small in scale and challenging to segment, such as gallbladder (Gall: 5.26\% improvement), stomach (Sto: 6.60\% improvement), pancreas (Pan: 7.58\% improvement), etc., the performance improvements highlight MambaMIM's capability to effectively capture both local and global representations.

\begin{figure}[!t]
    \centering
    \includegraphics[width=0.48\textwidth]{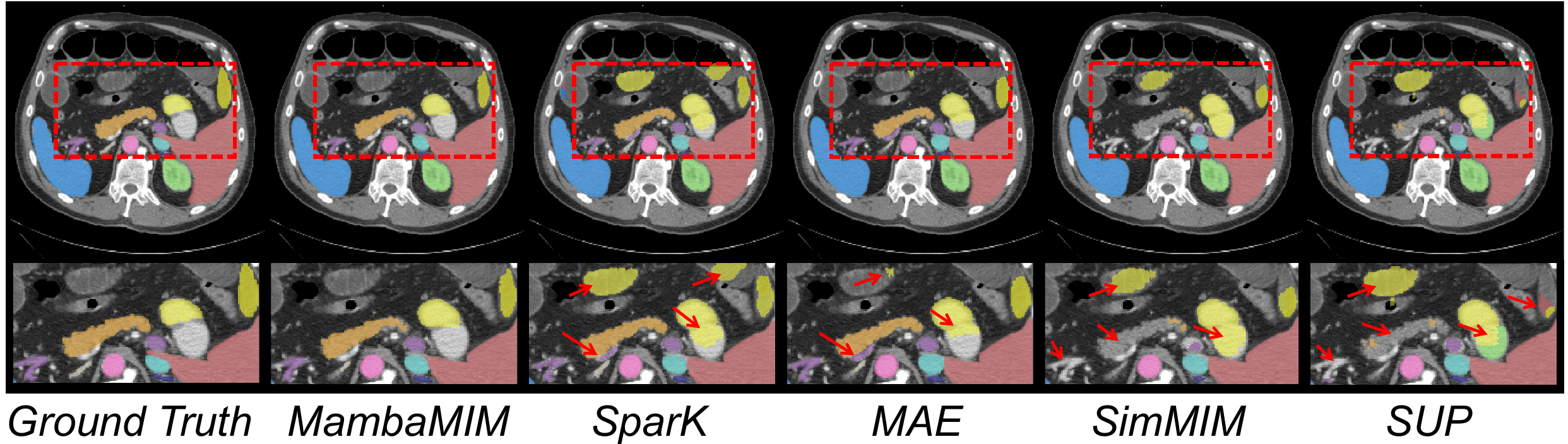}
    \vspace{-6mm}
    \caption{Visualization results on BTCV dataset.}
    \label{fig:btcv}
    \vspace{-4mm}
\end{figure}

\begin{figure}[!t]
    \centering
    \includegraphics[width=0.48\textwidth]{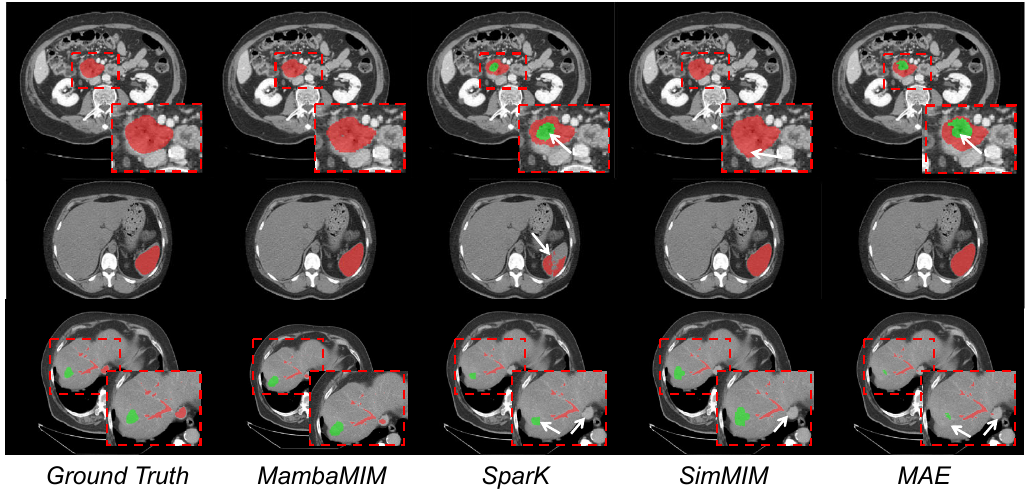}
    \vspace{-6mm}
    \caption{Visualization results on MSD datasets.}
    \vspace{-4mm}
    \label{fig:msd}
\end{figure}

\renewcommand{\multirowsetup}{\centering}  
\begin{table*}[!]
\centering
\caption{Best performance of other state-of-the-art pre-training methods on AMOS. \textbf{val} (bold) / \underline{val} (underline) : top method / second method. $\dagger$ denotes we utilize official pre-training weights. { The \textit{P}-value indicates the one-sample \textit{t}-test results.}\label{tab.AMOS}}
\vspace{-2mm}
\resizebox{1\linewidth}{!}
{
\begin{tabular}{l | l | ccccccccccccccc | c | l }
\Xhline{1px} 
\multicolumn{2}{c|}{Pretrain Method} & \multicolumn{15}{c|}{AMOS}  & \multirow{2}{*}{Avg (\%)} & \multirow{2}{*}{{\textit{P}-value}} \\
\cline{1-17}
Method & Network & Spl & R-kid & L-kid & Gall & Eso & Liver & Sto & Aor & Pos & Pan & R-AG & L-AG & Duo & Bla & Pro & & \\
\hline

MG$^{\dagger}$ & 3D U-Net & 93.84 & 91.51 & 91.93 & 75.30 & 0.00 & 93.82 & 88.45 & 93.25 & 88.22 & 83.62 & 0.00 & 0.00 & 74.57 & 70.27 & 0.00 & 62.98 & {$\leq 0.0001$} \\

TransVW$^{\dagger}$ & 3D U-Net  & 94.01 & 92.12 & 92.20 & 74.32 & 80.36 & 95.38 & 88.71 & 93.24 & 88.35 & 83.65 & 72.19 & 72.72 & 76.07 & 70.46 & 64.89 & 82.57 & {$\leq 0.0001$} \\

MAE & UNETR  & \textbf{95.94} & \underline{93.83} & \underline{95.24} & 75.57 & 80.03 & \textbf{97.16} & 90.44 & 93.54 & 88.29 & 84.33 & 72.19 & 71.66 & 76.97 & \underline{72.56} & 66.54 & 83.61 & {$\leq 0.05$} \\

UniMiSS$^{\dagger}$ & MiT & 94.20 & 92.41 & 93.84 & 71.77 & 72.73 & 96.09 & 86.64 & 91.43 & 84.63 & 79.17 & 66.53 & 67.28 & 70.49 & 68.90 & 62.64 & 79.91 & {$\leq 0.0001$} \\

SUP$^{\dagger}$ & Swin UNETR & 95.07 & 93.40 & 94.36 & 75.79 & 79.23 & 96.69 & 88.13 & 93.05 & 87.49 & 82.29 & 71.25 & 70.78 & 74.64 & 69.14 & 65.49 & 82.45 & {$\leq 0.0001$} \\

SparK & MedNeXt & 95.65 & 93.71 & \textbf{95.24} & \textbf{78.51} & 80.98 & 97.01 & \textbf{91.30} & 93.72 & 88.93 & 84.65 & 71.66 & 73.59 & 77.92 & \textbf{72.71} & 65.59 & 84.07 &  {$> 0.05$} \\

PCRLv2$^{\dagger}$ & 3D U-Net & 89.21 & 87.69 & 88.14 & 67.84 & 0.00 & 93.97 & 81.08 & 91.24 & 82.02 & 71.65 & 0.00 & 0.00 & 59.29 & 0.00 & 0.00 & 54.14 & {$\leq 0.0001$} \\

GVSL$^{\dagger}$ & 3D U-Net & 94.22 & 92.47 & 93.26 & 73.95 & 78.63 & 96.17 & 87.85 & 92.36 & 86.24 & 81.35 & 70.25 & 69.97 & 73.11 & 67.92 & 63.00 & 81.38 & {$\leq 0.0001$} \\

vox2vec$^{\dagger}$ & 3D U-Net  & 88.34 & 89.80 & 87.54 & 66.51 & 69.39 & 93.63 & 77.95 & 90.46 & 81.42 & 70.69 & 63.82 & 60.58 & 60.71 & 64.98 & 55.85 & 74.77 & {$\leq 0.0001$} \\

VoCo$^{\dagger}$ & Swin UNETR & 95.57 & \textbf{94.08} & 95.07 & 78.43 & \underline{81.78} & 96.89 & \underline{90.67} & \textbf{93.93} & \underline{89.25} & \underline{85.37} & \underline{73.49} & \underline{74.43} & \textbf{78.53} & 72.14 & \textbf{66.91} & \underline{84.43}  & {$> 0.05$} \\

\rowcolor{gray!15} MambaMIM & HyMamba & \underline{95.91} & 93.65 & 94.66 & \underline{78.45} & \textbf{82.00} & \underline{97.10} & 90.60 & \underline{93.87} & \textbf{89.37} & \textbf{85.44} & \textbf{74.39} & \textbf{75.29} & \underline{78.18} & 71.82 & \underline{66.88} & \textbf{84.50}  & \multicolumn{1}{c}{-} \\

\Xhline{1px} 

\end{tabular}
}
\vspace{-2mm}
\end{table*}

\renewcommand{\multirowsetup}{\centering}  
\begin{table*}[!]
\centering

\caption{{Result on unseen datasets of CT-ORG and KiTS 23.} \textbf{{val}}{ (bold) / }\underline{{val} }{(underline) : top method / second method. $\dagger$ denotes we utilize official pre-training weights. The \textit{P}-value indicates the one-sample \textit{t}-test results.}\label{tab:unseen} }
\vspace{-2mm}
\resizebox{1\linewidth}{!}
{
\begin{tabular}{l | l | ccccccl | ccccl | c}
\Xhline{1px} 

\multicolumn{2}{c|}{{Pre-training method}} 
& \multicolumn{7}{c|}{{CT-ORG}}
& \multicolumn{5}{c|}{{KiTS 23}}
&  \multirow{2}{*}{{Avg (\%)}}
\\
\cline{1-14}
\multicolumn{1}{c|}{{Method}} & \multicolumn{1}{c|}{{Network}}  & \multicolumn{1}{c}{{Liver}} & \multicolumn{1}{c}{{Bladder}} & \multicolumn{1}{c}{{Lungs}} & \multicolumn{1}{c}{{Kidneys}} & \multicolumn{1}{c}{{Bone}} & \multicolumn{1}{c}{{Avg (\%)}} & \multicolumn{1}{c|}{{\textit{P}-value}} & \multicolumn{1}{c}{{kidney}} & \multirow{1}{*}{{Tumor}} & \multirow{1}{*}{{Cyst}} & \multicolumn{1}{c}{{Avg (\%)}} & \multicolumn{1}{c|}{{\textit{P}-value}} \\
\hline

{MG$^{\dagger}$} & {3D U-Net} & {94.65} & {65.39} & {96.97} & {90.23} & {88.86} & {87.22} & {$\leq 0.01$} & {92.89} & {63.82} & {0.0} & {52.23} & {$\leq 0.0001$} & {69.72} \\

{TransVW$^{\dagger}$} & {3D U-Net} & {94.80} & {61.25} & {96.45} & {90.75} & {88.96} & {86.44} & {$\leq 0.01$} & {92.70} & {66.47} & {26.87} & {62.02} & {$\leq 0.0001$} & {74.23} \\

{$\text{MAE}$} & {UNETR} & \underline{{96.16}} & \underline{{66.23}} & {97.36} & \textbf{{93.06}} & {90.62} & \underline{{88.68}} & {$> 0.05$} & {92.82} & {61.53} & {24.39} & {59.58} & {$\leq 0.0001$} & {74.13} \\

{UniMiSS$^{\dagger}$} & {MiT} & {94.46} & {65.78} & {96.59} & {90.45} & {89.42} & {87.34} & {$\leq 0.01$} & {91.84} & {64.71} & {23.66} & {60.07} & {$\leq 0.0001$} & {73.70} \\

{$\text{SUP}^{\dagger}$} & {Swin UNETR} & {94.99} & {63.16} & {97.19} & {91.06} & {90.39} & {87.35} & {$\leq 0.01$} & {92.76} & {66.26} & {27.16} & {62.06} & {$\leq 0.0001$} & {74.70} \\

{SparK} & {MedNeXt} & {95.99} & {67.83} & {97.25} & {90.15} & {90.28} & {88.30} & {$\leq 0.001$} & {93.30} & {71.82} & {29.49} & {64.87} & {$\leq 0.05$} & {76.58} \\

{PCRLv2$^{\dagger}$} & {3D U-Net} & {94.62} & {0.0} & {95.13} & {84.46} & {88.69} & {72.58} & {$\leq 0.0001$} & {89.42} & {59.68} & {0.0} & {49.70} & {$\leq 0.0001$} & {61.14} \\

{GVSL$^{\dagger}$} & {3D U-Net} & {95.04} & {66.68} & {97.11} & {88.89} & {90.36} & {87.61} & {$\leq 0.01$} & {92.72} & {68.41} & {27.53} & {62.89} & {$\leq 0.0001$} & {75.25} \\

{vox2vec$^{\dagger}$} & {3D U-Net} & {92.73} & {61.56} & {95.31} & {83.31} & {89.12} & {84.40} & {$\leq 0.01$} & {91.33} & {65.70} & {23.79} & {60.27} & {$\leq 0.0001$} & {72.33} \\

{$\text{VoCo}^{\dagger}$} & {Swin UNETR} & \textbf{{96.35}} & {65.75} & \underline{{97.50}} & {92.31} & \underline{{90.74}} & {88.53} & {$> 0.05$} &  \underline{{93.54}} & \underline{{72.00}} & \underline{{30.34}} & \underline{{65.29}} & {$\leq 0.05$} & \underline{{76.91}} \\

\rowcolor{gray!15} {$\text{MambaMIM}$} & {HyMamba} & {96.08} & \textbf{{69.97}} & \textbf{{97.71}} & \underline{{92.56}} & \textbf{{90.79}} & \textbf{{89.42}} & \multicolumn{1}{c|}{-} & \textbf{{93.70}} & \textbf{{75.34}} & \textbf{{32.53}} & \textbf{{67.19}} & \multicolumn{1}{c|}{-} & \textbf{{78.30}} \\
\Xhline{1px}

\end{tabular}
}
\vspace{-2mm}
\end{table*}

\renewcommand{\multirowsetup}{\centering}  
\begin{table}[!]
\centering
\caption{Experimental results on BraTS 21. TC, WT, and ET denote the tumor core, whole tumor, and enhancing tumor, respectively. \textbf{val} (bold) / \underline{val} (underline) : top method / second method. $\dagger$ denotes we utilize official pre-training weights.}
\vspace{-2mm}
\resizebox{1\linewidth}{!}
{
\begin{tabular}{ l | l | ccc | c}
\Xhline{1px} 
\multirow{1}{*}{Pretrain Method} & \multirow{1}{*}{\# Backbone} & TC & WT & ET  & \multirow{1}{*}{Avg (\%)} \\
\hline

SparK & MedNeXt & \underline{89.37} & 92.02 & 84.80 & 88.73 \\
UniMiSS$^{\dagger}$  & MiT & 87.58 & 91.32 & 82.66 & 87.18 \\
MG$^{\dagger}$ & 3D U-Net & 88.56 & 92.34 & 83.99 & 88.30 \\
TransVW$^{\dagger}$ & 3D U-Net & 86.99 & 91.88 & 70.76 & 83.21 \\
GVSL$^{\dagger}$ & 3D U-Net & 88.16 & 91.27 & 83.42 & 87.62 \\
vox2vec$^{\dagger}$ & 3D U-Net & 74.19 & 83.01 & 70.12 & 75.77 \\
SUP$^{\dagger}$ & Swin UNETR & 89.22 & \underline{92.51} & \underline{85.60} & \underline{89.11} \\
VoCo$^{\dagger}$ & Swin UNETR & 89.21 & 91.93 & 85.28 & 88.81 \\
MAE & UNETR & 87.98 & 92.01 & 84.66 & 88.22 \\
\rowcolor{gray!15} MambaMIM & HyMamba & \textbf{90.38} & \textbf{92.87} & \textbf{86.48} & \textbf{89.91} \\

\Xhline{1px} 

\end{tabular}
}

\vspace{-2mm}
\label{tab:brats}
\end{table}

{Since different networks have different optimal pre-training methods, we select the general and medical SSL methods with the best pre-training performance for each network shown in Table~\ref{tab:msd} and Table~\ref{tab.AMOS}. Our proposed MambaMIM outperforms other pre-training methods. In Table~\ref{tab:msd}, MambaMIM achieves improvements of at least 1.59\% (64.96\% vs 63.37\%) in Pancreas and 0.49\% (64.91\% vs 64.42\%) in Hepatic Vessel, due to our proposed TOKI. TOKI further enhances Mamba's global localization capability. For multi-organ segmentation tasks, MambaMIM also gains the highest performance with 80.16\% DSC in BTCV and 84.50\% DSC in AMOS. Visualization results on lesion and abdomen are shown in Fig.~\ref{fig:msd} and Fig.~\ref{fig:btcv}. It can be seen that MambaMIM enables precise lesion and multi-scale organ segmentation results.}

\noindent\textbf{Performance of different pre-training methods on the same baseline model.} We also validate different pre-training methods on BTCV with the same backbone (HyMamba). The average DSC scores on the BTCV datasets for all contrastive-based pre-training methods are presented in Table~\ref{tab:btcv3}. MambaMIM significantly outperforms other self-supervised methods, achieving a minimum improvement of 1.88\% (80.16\% vs 78.28\%) in segmentation performance. { Our empirical findings suggest that masked image modeling approaches like MambaMIM offer more effective transfer gains for vision Mamba architectures in medical image segmentation tasks. This aligns with our understanding that contrastive approaches (\textit{i.e.}, MoCov2, SimSiam, SUP, SwAV, and VoCo) primarily learn \textbf{high-level semantic representations} suitable for \textbf{classification} tasks~{\citep{highlevel}}, which may not fully align with the requirements of dense prediction tasks like segmentation that demand localized feature learning. In contrast, MIM-based methods, such as MambaMIM, have been shown to excel at capturing fine-grained, localized features by reconstructing masked regions, thereby learning detailed spatial dependencies and local structures~{\citep{lowlevel}}.}

\renewcommand{\multirowsetup}{\centering}  
\begin{table*}[!h]
\centering
\caption{Ablation study on mask ratio for the BTCV 3D segmentation task. \textbf{val} (bold) / \underline{val} (underline) : top method / second method.\label{tab:ablation1}}
\vspace{-2mm}
\resizebox{1\linewidth}{!}
{
\begin{tabular}{l | ccccccccccc | c}
\Xhline{1px} 
\multicolumn{1}{l |}{Mask Ratio}  & \multirow{1}{*}{Spl} & \multirow{1}{*}{Kid} & \multirow{1}{*}{Gall} & \multirow{1}{*}{Eso} & \multirow{1}{*}{Liv} & \multirow{1}{*}{Sto} & \multirow{1}{*}{Aor} & \multirow{1}{*}{IVC} & \multirow{1}{*}{Veins} & \multirow{1}{*}{Pan} & \multirow{1}{*}{AG} & \multirow{1}{*}{Avg (\%)} \\
\hline
w/o pretrained & 88.60 & 87.18 & 60.67 & 74.73 & 94.34 & 79.60 & 88.67 & 82.07 & 68.83 & 71.53 & 62.43 & 77.56 \\
mask 25 \% & 89.18 & 87.75 & 63.93 & \underline{74.92} & 94.81 & 83.99 & 89.71 & 83.40 & 69.42 & \underline{77.26} & 64.48 & 79.31 \\

mask 50 \% & \textbf{90.75} & 87.72 & 64.06 & 74.76 & 94.63 & 84.17 & \underline{90.54} & \underline{84.25} & \textbf{72.01} & 75.30 & \textbf{65.60} & \underline{79.78} \\
mask 75 \% & \underline{90.47} & \textbf{88.06} & \textbf{65.93} & 73.91 & \underline{95.00} & \textbf{86.20} & \textbf{91.08} & \textbf{84.26} & \underline{70.40} & \textbf{79.11} & \underline{64.82} & \textbf{80.16} \\

{mask 80  \%} & {89.52} & \underline{{87.77}} & 64.58 & {\textbf{75.37}} & {\textbf{95.05}} & \underline{{84.61}} & {90.14} & {83.32} & {70.02} & {77.06} & {64.59} & {79.57} \\

{mask 90  \%} & {89.85} & {87.02} & \underline{{65.34}} & {73.95} & {94.79} & {83.80} & {90.51} & {83.58} & {69.65} & {76.11} & {63.62} & {79.14} \\

\Xhline{1px}
\end{tabular}
}
\vspace{-2mm}
\end{table*}

\renewcommand{\multirowsetup}{\centering}  
\begin{table}[t!]
\centering

\caption{Ablation on each component in MambaMIM. The experiment is performed on BTCV for the 3D segmentation task.\label{tab:ablation}}
\vspace{-2mm}
\resizebox{1\linewidth}{!}
{
\begin{tabular}{l | c c c | c}
\Xhline{1px} 
\multirow{2}{*}{MambaMIM components}  & \multirow{1}{*}{Hybrid} & \multirow{1}{*}{Skip}  & \multirow{2}{*}{TOKI}  & \multirow{2}{*}{Avg (\%)}\\
  &  MIM & -connection &  &   \\
\hline
w/o pretrained &  — & — & — & 77.56 \\
w/o bottom-up masking & \ding{55} & \ding{55} & \ding{55} & 79.20 \\
w/o skip &  \checkmark & \ding{55}  & \ding{55} & 79.25 \\
w/o TOKI &  \checkmark &  \checkmark & \ding{55} &  79.38 \\
MambaMIM & \checkmark & \checkmark &  \checkmark & \textbf{80.16}\\

\Xhline{1px}
\end{tabular}
}
\vspace{-2mm}
\end{table}

\begin{figure}[!t]
    \centering
    \includegraphics[width=0.48\textwidth]{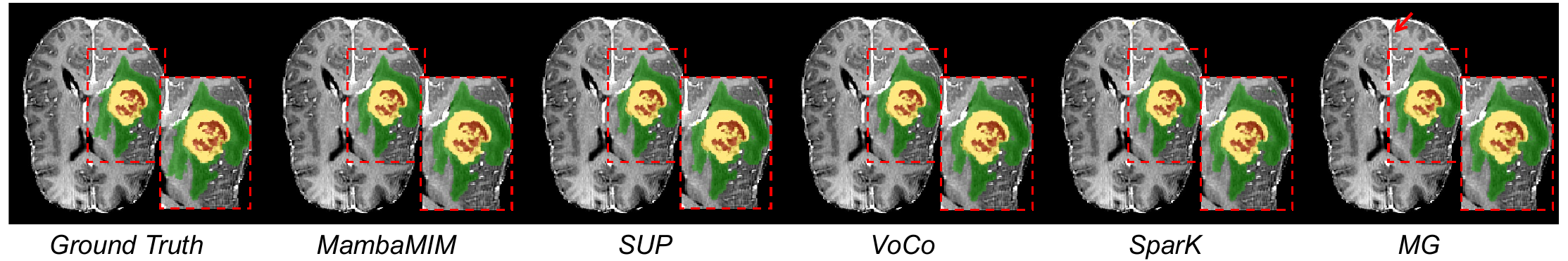}
    \vspace{-6mm}
    \caption{Visualization results on BraTS 21 datasets.}
    \label{fig:brats}
\end{figure}

\begin{figure}[!t]
    \centering
    \includegraphics[width=0.48\textwidth]{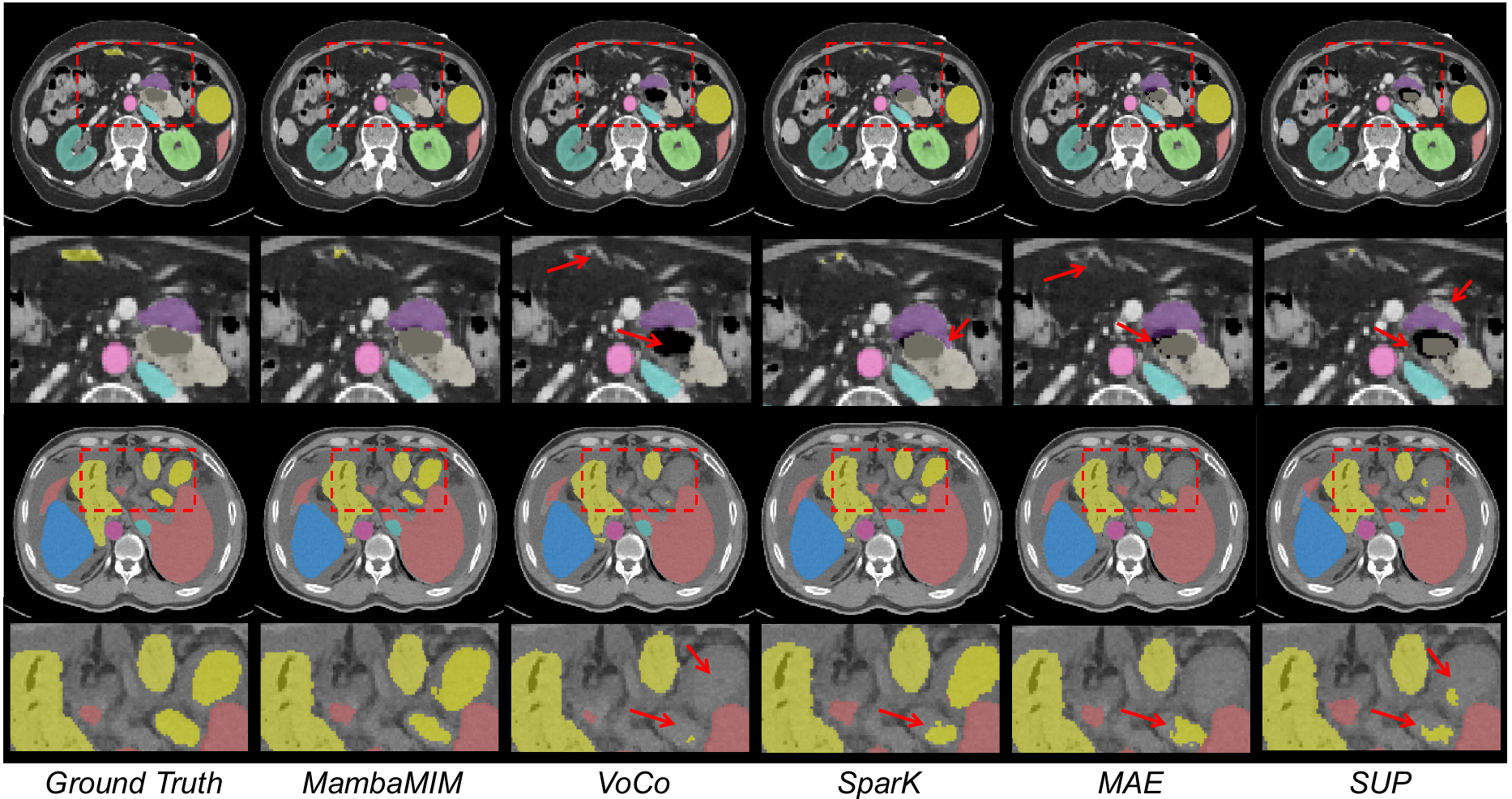}
    \vspace{-6mm}
    \caption{Visualization results on AMOS datasets.}
    \label{fig:amos}
\end{figure}

\noindent {\textbf{Promising performance on unseen datasets.} To further validate the generalizability of our approach, we conducted experiments on \textbf{unseen} datasets in pre-training, i.e., CT-ORG~{\citep{ctorg}} and KiTS 23~{\citep{kits21}}. The results on CT-ORG and KiTS 23 are shown in Table~{\ref{tab:unseen}}. It can be seen that MambaMIM achieves the best performance with 89.42\% and 67.19\% Dice Score on unseen datasets CT-ORG and KiTS 23, respectively. Notably, MambaMIM shows a 1.39\% improvement and attains an average Dice Score of 78.30\%, outperforming other well-known self-supervised methods. This highlights our MambaMIM's strong generalizability and robustness by showing superior performance across diverse and unseen datasets.}

\noindent\textbf{Generalization to MRI modalities.} 
Generalization is also an important area in medical image analysis due to the scarcity of medical images. Our results prove that the performance gain of MambaMIM is generalizable to other modalities. As shown in Table~\ref{tab:brats}, we evaluate the downstream on a widely used MRI dataset, \textit{i.e.} BraTS 21~\citep{brats21}. MambaMIM achieves the highest DSC score, \textit{i.e.} 89.91\%, surpassing existing SOTA methods. Despite the significant differences in high-level semantics between MRI and CT, this generalization ability highlights that MambaMIM effectively learns local representations that are independent of high-level semantics, thereby improving its performance across different modalities.

\begin{figure}[!t]
\centering
\begin{minipage}[t]{0.16\textwidth}
\centering
\includegraphics[width=2.8cm]{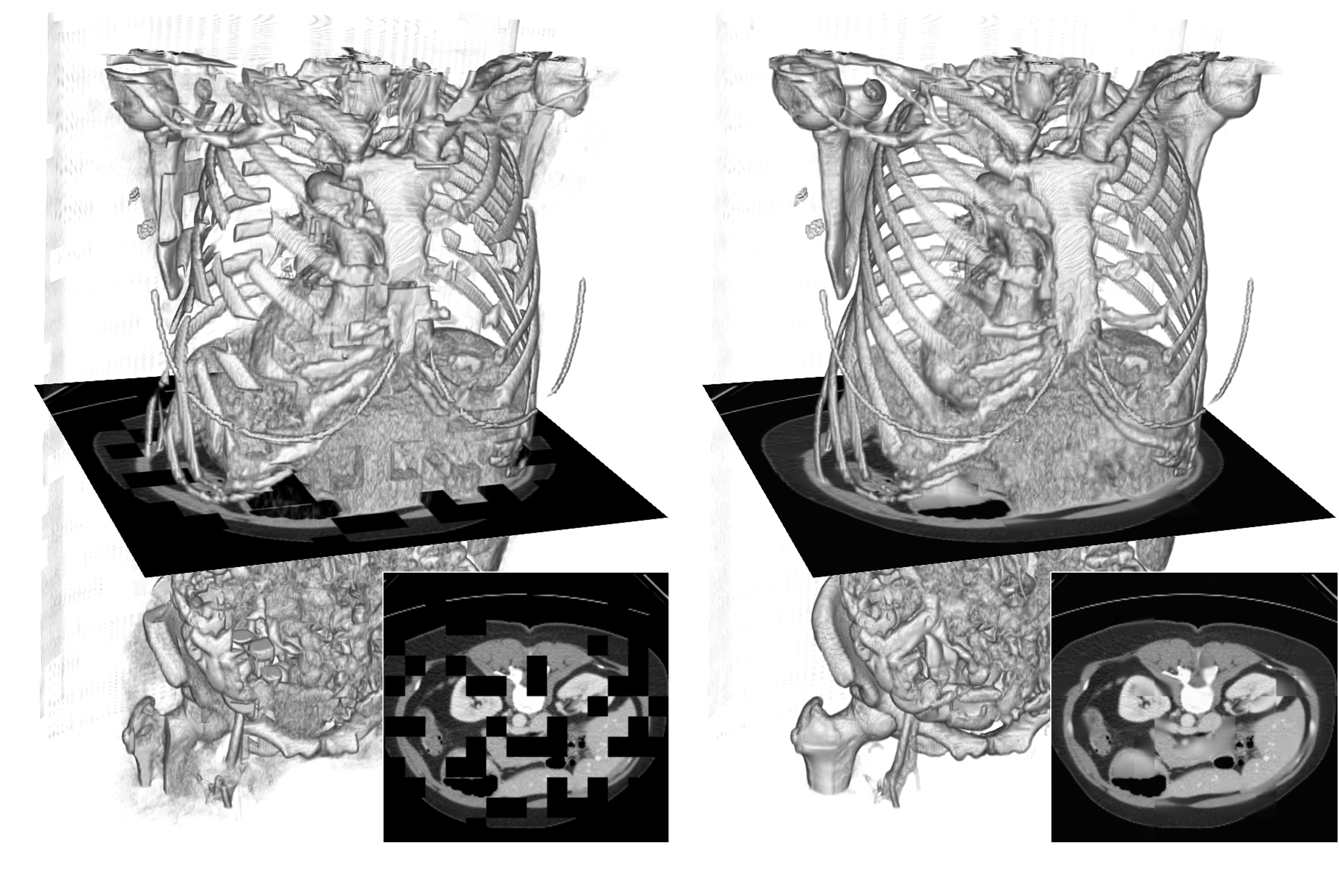}
\\ (a)
\end{minipage}
\hspace{-0.24cm}
\begin{minipage}[t]{0.16\textwidth}
\centering
\includegraphics[width=2.8cm]{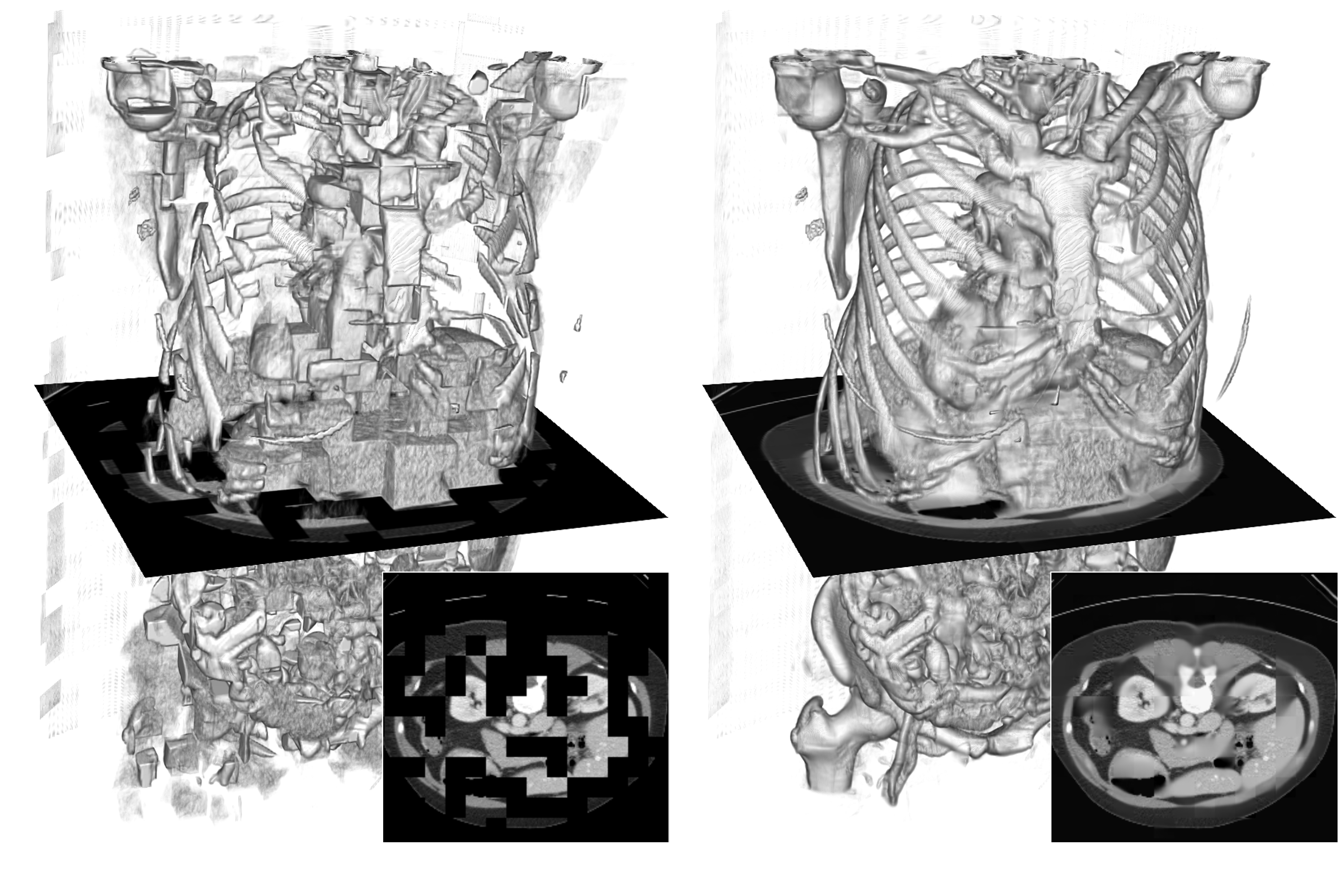}
\\ (b)
\end{minipage}
\hspace{-0.24cm}
\begin{minipage}[t]{0.16\textwidth}
\centering
\includegraphics[width=2.8cm]{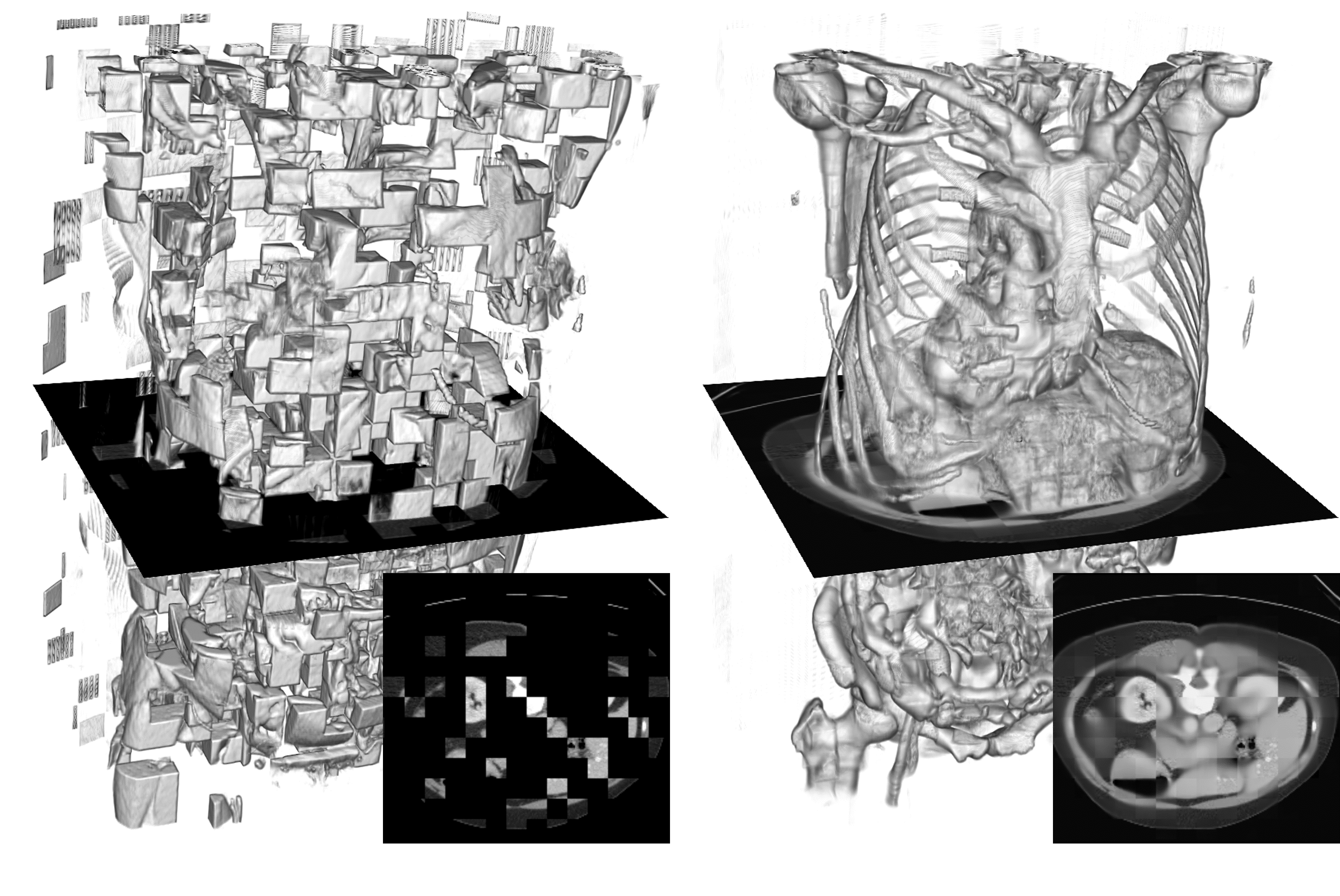}
\\ (c)
\end{minipage}
\caption{Reconstruction results by MambaMIM with different mask ratios (left: masked input, right: reconstruction result): (a) mask 25\% (b) mask 50\% (c) mask 75\%. \label{fig:re}}
\vspace{-4mm}
\end{figure}

\noindent\textbf{Visualization.} We visualize 3D reconstruction results with different mask ratios to show what MambamMIM learns in the pre-training. As shown in Fig.~\ref{fig:re}, our method can almost reconstruct the different shapes of organs, bones, and other details from the different portions of unmasked patches.

\subsection{Ablation study}

\noindent\textbf{Ablation study on the mask ratio.}
Table~\ref{tab:ablation1} shows the impact of different mask ratios on the model's performance. A relatively small 25\% mask ratio substantially boosts the performance from 77.56\% to 79.31\%, indicating that even a small amount of masking significantly enhances the performance, which demonstrates the effectiveness of generative pre-training. Increasing the mask ratio results in a consistent performance improvement. At a masking rate of 75\%, the segmentation performance of the downstream network reaches its peak at 80.16\%. This suggests that, within a certain threshold, masking a greater number of blocks allows the pre-trained encoder to more effectively encode the input into latent representations, a capability that is critical for optimizing downstream segmentation tasks.

\begin{figure}[!t]
    \centering
    \includegraphics[width=0.48\textwidth]{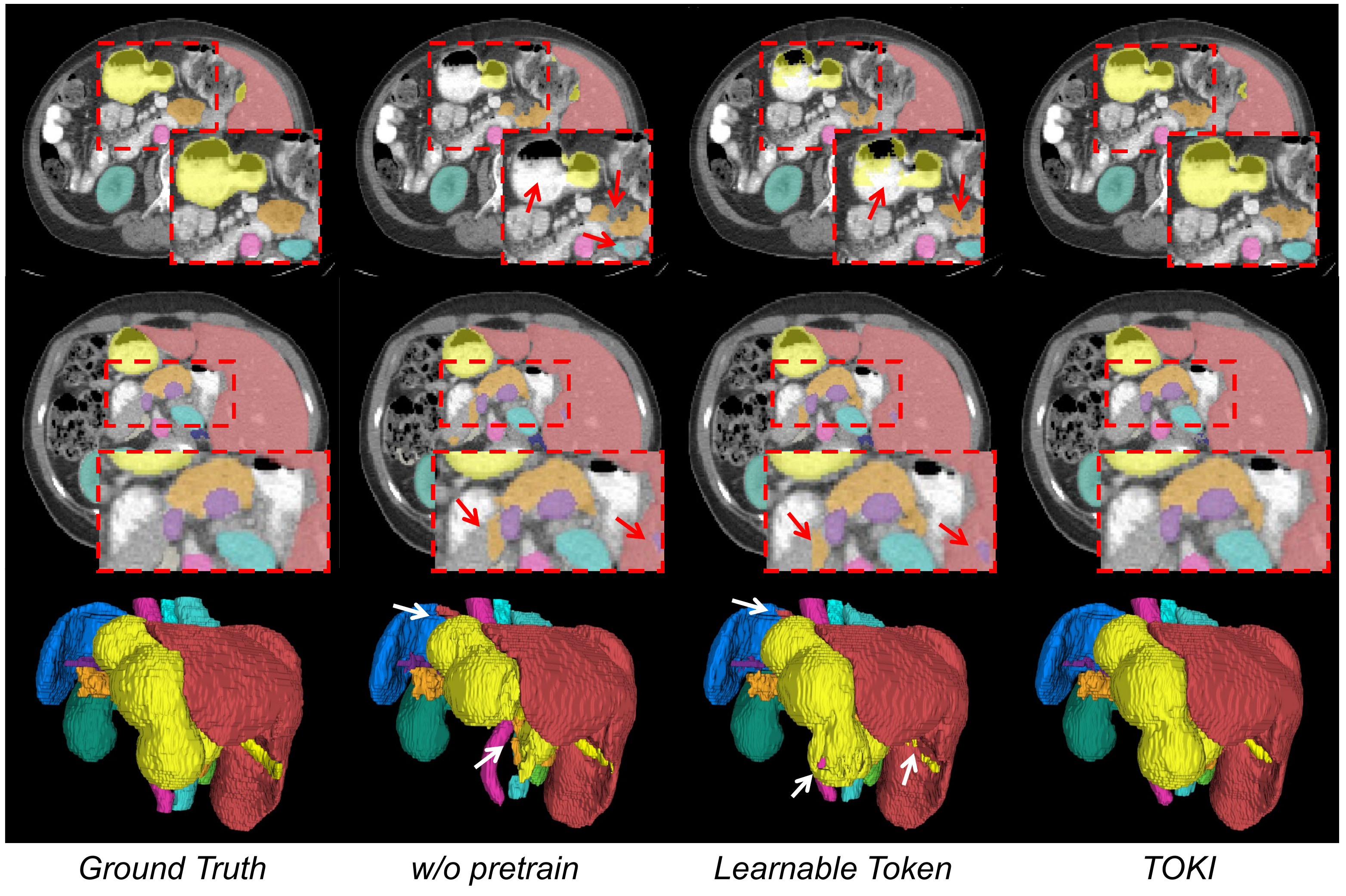}
    \vspace{-4mm}
    \caption{Visualization of ablation results on BTCV dataset.}
    \label{fig:ablation}
    \vspace{-4mm}
\end{figure}

\noindent\textbf{Ablation on MambaMIM components.}
Table~\ref{tab:ablation} explores the effects of omitting various components of the MambaMIM framework. Removing the bottom-up masking leads to a performance drop, showing that bottom-up masking contributes to the model's performance. Although the upper CNN accounts for the majority of the hybrid model's performance, pre-training still greatly boosts the performance without a consistent mask. Nevertheless, our bottom-up masking strategy still further lifts the performance. 
Incorporating skip-connection in pre-training improves the performance from 79.25\% to 79.38\%,  which demonstrates the importance of upstream and downstream pattern alignment. Utilizing our proposed TOKI instead of learnable token greatly enhances the performance from 79.38\% to 80.16\%, highlighting the effectiveness of state-space interpolation in Mamba pre-training. 
Additionally, we visualize the segmentation results under different ablation components. As in Fig.~\ref{fig:ablation}, our initiative demonstrates significant benefits for downstream tasks.

\renewcommand{\multirowsetup}{\centering}  
\begin{table*}[h!]
\centering
\caption{{Results of different scanning method on BTCV.} \textbf{{val}}{ (bold) / }\underline{{val} }{(underline) : top method / second method.}\label{tab:scanbtcv}.}
\vspace{-2mm}
\resizebox{1\linewidth}{!}
{
\begin{tabular}{l | c | ccccccccccc| c}
\Xhline{1px}
\multicolumn{1}{c |}{{Network}} & \multicolumn{1}{c |}{{Scanning}}  & \multirow{1}{*}{{Spl}} & \multirow{1}{*}{{Kid}} & \multirow{1}{*}{{Gall}} & \multirow{1}{*}{{Eso}} & \multirow{1}{*}{{Liv}} & \multirow{1}{*}{{Sto}} & \multirow{1}{*}{{Aor}} & \multirow{1}{*}{{IVC}} & \multirow{1}{*}{{Veins}} & \multirow{1}{*}{{Pan}} & \multirow{1}{*}{{AG}} & \multirow{1}{*}{{Avg (\%)}} \\
\hline
 \multirow{4}{*}{{HyMamba (MambaMIM)}} & {Raster}  & {89.67} & {87.23} & \underline{{65.93}} & {73.91} & \underline{{95.00}} & {\textbf{86.20}} & {\textbf{91.08}} & \underline{{84.26}} & {70.40} & {\textbf{79.11}} & {64.81} & {80.16} \\ 
 
 & {Zigzag} & \underline{{90.11}} & {88.04} & {65.71} & {73.71} & {94.75} & {84.34} & {89.92} & {84.10} & {71.33} & {77.23} & \underline{{66.87}} & {80.08} \\ 

 & {Hilbert} & {\textbf{90.45}} & {\textbf{88.94}} & {\textbf{66.34}} & \underline{{74.31}} & {\textbf{95.12}} & {85.44} & \underline{{90.55}} & {83.51} & \underline{{71.33}} & \underline{{77.86}} & {65.83} & \underline{{80.34}} \\ 

 & {Shuffle} & {90.04} & \underline{{88.64}} & {65.26} & {\textbf{75.80}} & {94.98} & \underline{{85.86}} & {89.99} & {\textbf{84.67}} & {\textbf{73.26}} & {77.78} & {\textbf{68.36}} & {\textbf{80.89}} \\

\Xhline{1px}
\end{tabular}
}
\end{table*}

\begin{figure*}
    \centering
    \includegraphics[width=0.96\linewidth]{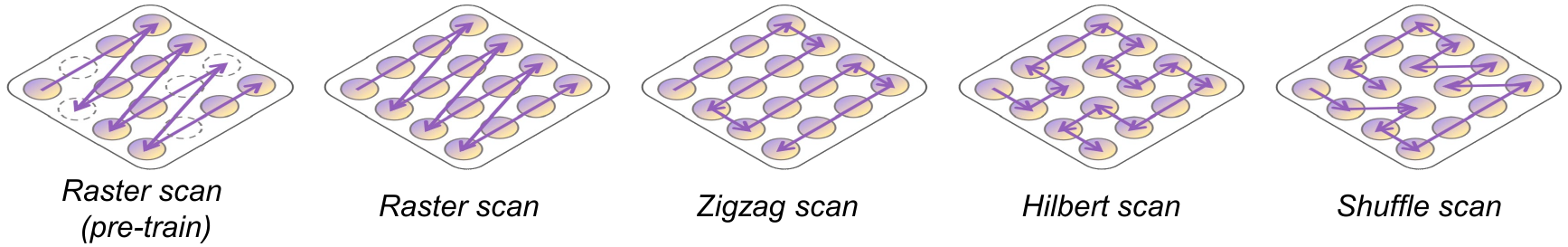}
    \caption{{The illustration of the different scanning methods.}}
    \label{fig:scanmethods}
\end{figure*}

\noindent {\textbf{Ablation on the scanning methods.} Table.~{\ref{tab:scanbtcv}} examines how scanning methods differ between pre-training (using standard raster scanning) and downstream impacts model performance, with evaluation across four prevalent scanning paradigms (shown in Fig.~{\ref{fig:scanmethods}}): Raster scan~{\citep{mamba, vim, localmamba}}, Zigzag scan~{\citep{yang2024plainmamba, groupmamba}}, Hilbert scan~{\citep{he2024mambaad, fractalmamba}}, and Shuffle scan~{\citep{zhu2024rethinking, shuffle}}. The experimental results indicate that all four evaluated scanning strategies demonstrate competitive performance, with the Shuffle scan achieving the highest fine-tuning performance 
 (80.89\% Dice score). This suggests that the pre-trained model exhibits robustness to variations in scan protocols. This finding suggests that downstream tasks could potentially employ various scanning strategies without significantly compromising the effectiveness of pre-trained models}

\section{Conclusions}
The success of Mamba in vision tasks prompts us to explore their potential in downstream tasks after being well pre-trained using large-scale unlabeled medical images. In this paper, we introduce MambaMIM, a novel generative self-supervised method based on selective structure state space sequence token-interpolation for single and hybrid Mamba architectures. Our method uses a bottom-up masking strategy to guarantee the consistency of masking between CNN and Mamba. Additionally, TOKI is employed to learn causal relationships between the masked sequences in the state space. Extensive downstream and ablation experiments demonstrate the superior performance of our pre-training method. We believe that MambaMIM can further benefit our community and our findings can inspire more work to maximize the potential of Mamba in various visual tasks.

\section{Acknowledgments}
Supported by Natural Science Foundation of China under
Grant 62271465, 62376153, 62402318, 24Z990200676, Suzhou Basic Research Program under Grant. SYG202338, Open Fund Project of Guangdong Academy of
Medical Sciences, China (No. YKYKF202206)

\bibliographystyle{model2-names.bst}\biboptions{authoryear}
\bibliography{refs}

\end{document}